\documentclass[sigconf]{acmart}
\AtBeginDocument{%
  \providecommand\BibTeX{{%
    \normalfont B\kern-0.5em{\scshape i\kern-0.25em b}\kern-0.8em\TeX}}}

\copyrightyear{2024} 
\acmYear{2024} 
\setcopyright{acmlicensed}\acmConference[KDD '24]{Proceedings of the 30th ACM SIGKDD Conference on Knowledge Discovery and Data Mining}{August 25--29, 2024}{Barcelona, Spain}
\acmBooktitle{Proceedings of the 30th ACM SIGKDD Conference on Knowledge Discovery and Data Mining (KDD '24), August 25--29, 2024, Barcelona, Spain}
\acmDOI{10.1145/3637528.3671964}
\acmISBN{979-8-4007-0490-1/24/08}

\usepackage{xcolor}
\definecolor{customgreen}{rgb}{0.5, 0.7, 0.5}
\usepackage{balance}
\begin{document}

\title{Team up GBDTs and DNNs: Advancing Efficient and Effective Tabular Prediction with Tree-hybrid MLPs}

\author{Jiahuan Yan}
\affiliation{%
  \institution{Zhejiang University}
  \city{Hangzhou}
  \country{China}}
\email{jyansir@zju.edu.cn}

\author{Jintai Chen$^*$}
\thanks{$^*$ The corresponding author.}
\affiliation{%
  \institution{University of Illinois at Urbana-Champaign}
  \city{Urbana}
  \state{IL}
  \country{USA}}
\email{jtchen721@gmail.com}

\author{Qianxing Wang}
\affiliation{%
  \institution{Zhejiang University}
  \city{Hangzhou}
  \country{China}}
\email{w.qianxing@zju.edu.cn}

\author{Danny Z. Chen}
\affiliation{%
  \institution{University of Notre Dame}
  \city{Notre Dame}
  \state{IN}
  \country{USA}}
\email{dchen@nd.edu}

\author{Jian Wu}
\affiliation{%
  \institution{Zhejiang University}
  \city{Hangzhou}
  \country{China}}
\email{wujian2000@zju.edu.cn}

\renewcommand{\shortauthors}{Jiahuan Yan, Jintai Chen, Qianxing Wang, Danny Z. Chen, \& Jian Wu}

% A conflict is a disagreement between at least two people or entities!! A contradiction is a disagreement within one person or 1 entity!!
\begin{abstract}
Tabular datasets play a crucial role in various applications. Thus, developing efficient, effective, and widely compatible prediction algorithms for tabular data is important. 
Currently, two prominent model types, Gradient Boosted Decision Trees (GBDTs) and Deep Neural Networks (DNNs), have demonstrated performance advantages on distinct tabular prediction tasks. However, selecting an effective model for a specific tabular dataset is challenging, often demanding time-consuming hyperparameter tuning. To address this model selection dilemma, this paper proposes a new framework that amalgamates the advantages of both GBDTs and DNNs, resulting in a DNN algorithm that is as efficient as GBDTs and is competitively effective regardless of dataset preferences for GBDTs or DNNs. 
Our idea is rooted in an observation that deep learning (DL) offers a larger parameter space that can represent a well-performing GBDT model, yet the current back-propagation optimizer struggles to efficiently discover such optimal functionality. On the other hand, during GBDT development, hard tree pruning, entropy-driven feature gate, and model ensemble have proved to be more adaptable to tabular data.
By combining these key components, we present a \textbf{T}ree-hybrid simple \textbf{MLP} (T-MLP). 
In our framework, a tensorized, rapidly trained GBDT feature gate, a DNN architecture pruning approach, as well as a vanilla back-propagation optimizer collaboratively train a randomly initialized MLP model. Comprehensive experiments show that T-MLP is competitive with extensively tuned DNNs and GBDTs in their dominating tabular benchmarks (88 datasets) respectively, all achieved with compact model storage and significantly reduced training duration. The codes and full experiment results are available at \url{https://github.com/jyansir/tmlp}.
\end{abstract}





\maketitle

\section{Introduction}
Tabular data are a ubiquitous and dominating data structure in various machine learning applications (e.g., click-through rate (CTR) prediction~\cite{covington2016deep} and financial risk detection~\cite{aziz2022machine}).
Current prevalent tabular prediction (i.e., classification and regression) models can be generally categorized into two main types: (1) Gradient Boosted Decision Trees (GBDTs)~\cite{friedman2001greedy,chen2016xgboost,ke2017lightgbm,prokhorenkova2018catboost}, a kind of classical non-deep-learning approach that has been extensively verified as test-of-time solutions~\cite{uddin2019comparing,borisov2022deep,grinsztajn2022tree}; (2) Deep Neural Networks (DNNs), on which continuous endeavors apply deep learning (DL) techniques from computer vision (CV) and natural language processing (NLP) to develop tabular learning methods such as meticulous architecture engineering~\cite{popov2019neural,arik2021tabnet,gorishniy2021revisiting,yan2023t2g,chen2022tabcaps} and pre-training~\cite{somepalli2022saint,wang2022transtab,zhu2023xtab}. With recent developments of bespoke tabular DNNs, increasing studies reported their better comparability~\cite{chen2022danets,yan2023t2g} and even superiority~\cite{somepalli2022saint,chen2023excelformer} to GBDTs, especially in complex data scenarios~\cite{wang2022transtab,ruiz2023enabling}, while classical thinking believes that GBDTs still completely surpass DNNs in typical tabular tasks~\cite{borisov2022deep,grinsztajn2022tree}, both evaluated with different benchmarks and baselines, implying respective tabular data proficiency of these two model types. 

For DNNs, their inherent high-dimensional feature spaces and smooth back-propagation optimization gain tremendous success on unstructured data~\cite{brown2020language,radford2021learning} and capability of mining subtle feature interactions~\cite{seo2017interpretable,song2019autoint,wang2021dcn,yan2023t2g}. 
Besides, leveraging DNN's transferability, recent popular tabular Transformers can be further improved by costly pre-training~\cite{somepalli2022saint,wang2022transtab,zhu2023xtab}, like their counterparts in NLP~\cite{kenton2019bert,brown2020language,zhao2023survey}.
However, compared to the simple multi-layer perceptron (MLP) and GBDTs, Transformer architectures are more complicated and are prone to be \textit{over-parameterized, data-hungry, and increase processing latency}, especially those recent language-model-based architectures~\cite{borisov2022language,zhang2023generative}. Thus, they typically under-perform on tabular datasets that are potentially small-sized~\cite{grinsztajn2022tree}.

Regarding GBDTs, they thrive on greedy feature selection, tree pruning, and efficient ensemble, yielding remarkable performances and efficiency on the majority of tabular prediction applications~\cite{uddin2019comparing,shwartz2022tabular}. Yet, they are usually \textit{hyperparameter-sensitive~\cite{prokhorenkova2018catboost,yan2024making} and not well-suited in extreme tabular scenarios}, such as large-scale tables with intricate feature interactions~\cite{ruiz2023enabling}. Also, their inference latency increases markedly as the data scale grows~\cite{borisov2022deep}.

Besides, both the GBDT and DNN frameworks achieve respective state-of-the-art results with \textit{expensive training costs}, since \textit{heavy hyperparameter search} is required to achieve considerable performance. But, this is carbon-unfriendly and is not compatible in computation-limited or real-time applications, while not enough proactive efforts on economical tabular prediction have been made.

\begin{figure*}[h]
  \centering
  \includegraphics[width=\textwidth]{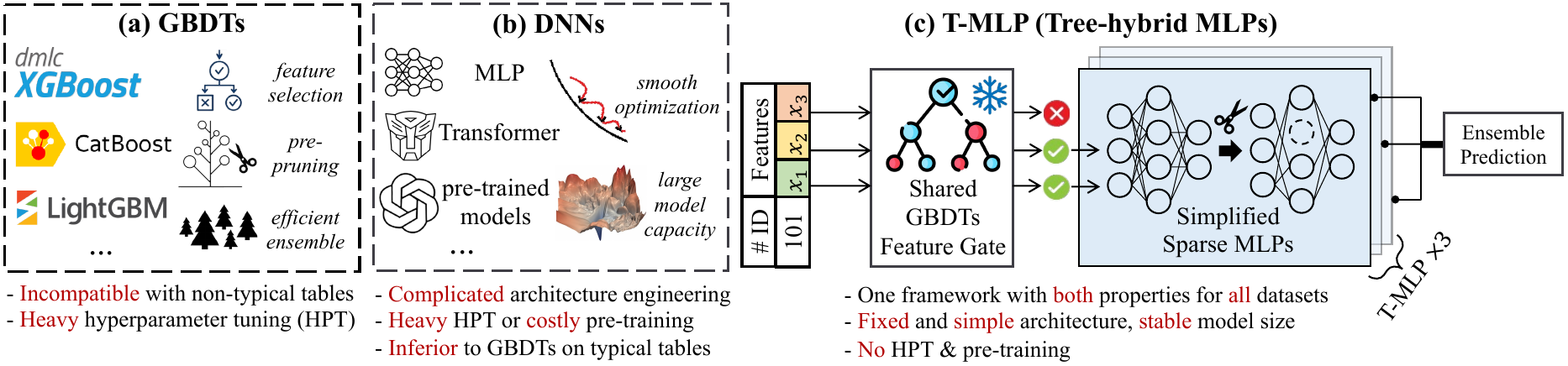}
  \caption{Our proposed T-MLP vs.~existing tabular prediction approaches: GBDTs and DNNs. (a) GBDTs are classical non-deep-learning models for tabular prediction. (b) DNNs are emerging promising methods especially for large-scale, complex, cross-table scenarios. (c) T-MLP is a hybrid framework that integrates the strengths of both GBDTs and DNNs, accomplished via GBDT feature gate tensorization, MLP framework pruning, simple block ensemble, and end-to-end back-propagation. 
  It yields competitive results on both DNN- and GBDT-favored datasets, with a rapid development process and compact model size.}
  \Description{Overall illustration for framework comparison of existing tabular prediction models and T-MLP. Current evaluations on GBDTs and DNNs give conflicting superiority with expensive training cost, while T-MLP with both properties from GBDTs and DNNs can produce satisfactory performances on all datasets under economical runtime cost and uniform configurations.}
  \label{fig:framework}
\end{figure*}

To address the model selection dilemma, we comprehensively combine the advantages of both GBDTs and DNNs, and propose a new \textbf{T}ree-hybrid simple \textbf{MLP} (T-MLP), which is high-performing, efficient, and lightweight.
Specifically, a single T-MLP is equipped with a GBDT feature gate to perform sample-specific feature selection in a greedy fashion, and GBDT-inspired pruned MLP architectures to process the selected salient features. 
The whole framework is optimized using back-propagation with these GBDTs' properties, and all the components make the system compact, overfit-resistant, and generalizable.
Furthermore, model ensemble can be efficiently achieved by training multiple sparse MLPs (we uniformly use 3 MLPs here) in parallel with a shared gate and predicting in a bagging manner. Overall, T-MLP has the following appealing features. (1) \textbf{Generalized data adaptability:} Different from existing tabular prediction methods that suffer from the model selection dilemma, T-MLP is flexible enough to handle all datasets regardless of the framework preference (see Sec.~\ref{main-exp} and Sec.~\ref{superiority-interpret}). (2) \textbf{Hyperparameter tuning free:} T-MLP is able to produce competitive results \textbf{with all the configurations pre-fixed}, which is significantly time-saving, user-friendly, environmentally friendly and widely practical in broader applications. (3) \textbf{Lightweight storage:} In T-MLP, the DNN part is purely composed of simple and highly sparse MLP architectures, yet is still able to be state-of-the-art competitive even with one-block MLP. Table~\ref{tab:cost-perform} presents the economical cost-performance trade-off of T-MLP compared to common DNNs; such cost-effectiveness becomes more profound as the data scale grows. 

In summary, our main contributions are as follows:

\renewcommand{\arraystretch}{0.8}
\begin{table}
\caption{Comparison of model cost-effectiveness on small and large datasets across popular tabular DNNs. $F$ and $N$ denote the amounts of features and samples,
%respectively, 
$P$ is the parameter number, and $T$ denotes the overhead of total training time against the proposed T-MLP. We reuse performances and parameter sizes of the best model configurations in the FT-Transformer benchmark. $T$ is evaluated on an NVIDIA A100 PCIe 40GB (see Sec.~\ref{exp-setup}). Based on the fixed architecture and training configurations, T-MLP achieves stable model size and cheap training duration cost regardless of the data scale.}
\vskip -0.8 em
\label{tab:cost-perform}
\centering
\begin{tabular}{@{}lcccccc@{}}
\toprule
\multicolumn{1}{c}{Dataset:} & \multicolumn{3}{c}{Adult ($F$=14, $N$=49K)}                   & \multicolumn{3}{c}{Year ($F$=90, $N$=515K)} \\ \midrule
\multicolumn{1}{l|}{}        & $P$(M) & $T$ & \multicolumn{1}{c|}{ACC $\uparrow$}   & $P$(M)  & $T$ & RMSE $\downarrow$ \\ \midrule
\multicolumn{1}{l|}{MLP}     & 0.77   & 7.7$\times$   & \multicolumn{1}{c|}{0.852} & 1.16    & 15.9$\times$   & 8.853 \\
\multicolumn{1}{l|}{NODE}    & 20.83   & 120.4$\times$ & \multicolumn{1}{c|}{0.858} & 7.55    & 206.0$\times$  & 8.784 \\
\multicolumn{1}{l|}{AutoInt} & 0.01   & 25.0$\times$  & \multicolumn{1}{c|}{0.859} & 0.08    & 101.9$\times$  & 8.882 \\
\multicolumn{1}{l|}{DCNv2}   & 1.18   & 8.0$\times$   & \multicolumn{1}{c|}{0.853} & 11.32   & 29.9$\times$   & 8.890 \\
\multicolumn{1}{l|}{FT-T}    & 3.82   & 19.6$\times$  & \multicolumn{1}{c|}{0.859} & 1.25    & 116.3$\times$  & 8.855 \\ \midrule
\multicolumn{1}{l|}{T-MLP}    & \textcolor{customgreen}{0.73}   & \textcolor{customgreen}{1.0$\times$}   & \multicolumn{1}{c|}{\textcolor{customgreen}{0.864}} & \textcolor{customgreen}{0.75}    & \textcolor{customgreen}{1.0$\times$}    & \textcolor{customgreen}{8.768} \\ \bottomrule
\end{tabular}
\vskip -0.8 em
\end{table}
\renewcommand{\arraystretch}{1}

\begin{itemize}
    \item We propose a new GBDT-DNN hybrid framework, T-MLP, which is a one-stop and economical solution for effective tabular data prediction regardless of framework preferences of specific datasets, offering a novel optimization paradigm for tabular model architectures.
    \item Multi-facet analysis on feature selection strategy, parameter sparsity, and decision boundary pattern is given for in-depth understanding of the T-MLP efficiency and superiority.
    \item Comprehensive experiments on 88 datasets from 4 benchmarks, covering DNN- and GBDT-favored ones, show that a single T-MLP is competitive with advanced or pre-trained DNNs, and T-MLP ensemble can even consistently outperform them and is competitive with extensively tuned state-of-the-art GBDTs, all achieved with a compact model size and significantly reduced training duration.
    \item We develop an open-source Python package with APIs of benchmark loading, uniform baseline invocation (DNNs, GBDTs, T-MLP), DNN pruning, and other advanced functions as a developmental tool for the tabular learning community.
\end{itemize}

\section{Related Work}

\subsection{Model Frameworks for Tabular Prediction}

In the past two decades, classical non-deep-learning methods~\cite{li1984classification,zhang2003learning,zhang2006learning,he2014practical} have been prevalent for tabular prediction applications, especially GBDTs~\cite{friedman2001greedy,chen2016xgboost,ke2017lightgbm,prokhorenkova2018catboost} due to their efficiency and robustness in typical tabular tasks~\cite{grinsztajn2022tree}.
Because of the universal success of DNNs on unstructured data and the development of computation devices, there is an increasing effort in applying DNNs to such tasks. The early tabular DNNs aimed to be comparable with GBDTs by emulating the ensemble tree frameworks (e.g., NODE~\cite{popov2019neural}, Net-DNF~\cite{katzir2020net}, and TabNet~\cite{arik2021tabnet}), but they neglected the advantages of DNNs for automatic feature fusion and interaction. Hence, more recent attempts leveraged DNNs' superiority, as they transferred successful neural architectures (e.g., AutoInt~\cite{song2019autoint}, FT-Transformer~\cite{gorishniy2021revisiting}), proposed bespoke designs (e.g., T2G-Former~\cite{yan2023t2g}), or adopted pre-training (e.g., SAINT~\cite{somepalli2022saint}, TransTab~\cite{wang2022transtab}), reporting competitive or even surpassing results compared to conventionally dominating GBDTs in specific data scenarios~\cite{somepalli2022saint,chen2023excelformer}. Contemporary surveys~\cite{borisov2022deep} demonstrated that GBDTs and DNNs are two prevailing types of frameworks in current tabular learning research.

\subsection{Lightweight DNNs}

Lightweight DNNs are an evergreen research topic in CV and NLP, which aim to maintain effective performance while promoting DNN compactness and efficiency. A recent trend is to substitute dominating backbones with pure simple MLPs, such as MLP-Mixer~\cite{tolstikhin2021mlp}, gMLP~\cite{liu2021pay}, MAXIM~\cite{tu2022maxim}, and other vision MLPs~\cite{guo2022hire,tang2022sparse,cao2023strip}, achieving comparable or even superior results to their CNN or Transformer counterparts with reduced capacity or FLOPs. This pure-MLP trend is also arising in NLP~\cite{fusco-etal-2023-pnlp} and other real-world applications~\cite{chen2023tsmixer}. Another lightweight scheme is model compression, where pruning is a predominant approach used to trim down large language models~\cite{vaswani2017attention,zhao2023survey} from various granularity~\cite{xia2022structured,ma2023llm,sun2023simple}. In the tabular prediction field, there are a few pure-MLP studies, but all focusing on regularization~\cite{kadra2021well} or numerical embedding~\cite{gorishniy2022embeddings} rather than the DNN architecture itself. Besides, model compression of tabular DNNs has not yet been explored. We introduce related techniques to make our T-MLP more compact and effective.

\section{Tree-hybrid Simple MLP}
We first review some preliminaries of typical GBDTs' inference process and feature encoding techniques in current Transformer-based tabular DNNs. Next, we elaborate on the detailed designs of several key components of T-MLP, including the GBDT feature gate for sample-specific feature selection, the pure-MLP basic block, and GBDT-inspired fine-grained pruning for sparse MLPs. Finally, we provide a discussion of the T-MLP workflow.

\subsection{Preliminaries}
\subsubsection*{\textbf{Problem Statement.}} Given a tabular dataset with input features $X \in \mathbb{R}^{N \times F}$ and targets $y \in \mathbb{R}^N$, the tabular prediction task is to find an optimal solution $f: \ \in \mathbb{R}^{N \times F} \to \mathbb{R}^N$ that minimizes the empirical difference between the predictions $\hat{y}$ and the targets $y$. Here in current practice, the common choice of $f$ is either traditional GBDTs (e.g., XGBoost~\cite{chen2016xgboost}, CatBoost~\cite{prokhorenkova2018catboost}, LightGBM~\cite{ke2017lightgbm}) or tabular DNNs (e.g., TabNet~\cite{arik2021tabnet}, FT-Transformer~\cite{gorishniy2021revisiting}, SAINT~\cite{somepalli2022saint}, T2G-Former~\cite{yan2023t2g}). A typical difference metric is accuracy or AUC score for classification tasks, and is the root of mean squared error (RMSE) for regression.

\subsubsection*{\textbf{Definition 3.1: GBDT Feature Frequency.}} Given a GBDT model with $T$ decision trees (e.g., CART~\cite{li1984classification}), the \textit{GBDT feature frequency} of a sample denotes the number of times each feature is accessed by this GBDT on the sample. Specifically, the process of GBDT inference on a sample $x \in \mathbb{R}^F$ provides $T$ times a single decision tree prediction $\hat{y}^{(k)} = \text{CART}^{(k)}(x), k \in \left \{ 1,2,\ldots,T \right \}$. For each decision tree prediction, there exists a sample-specific decision path from its root to one of the leaf nodes, forming a used feature list that includes features involved in this prediction action. We denote this accessed feature list of the $k$-th decision tree as a binary vector $\alpha^{(k)} \in \left \{0, 1\right \}^F$, in which 0 indicates that the corresponding feature of this sample is not used by the $k$-th decision, and 1 indicates that it is accessed. Consequently, we can represent the \textit{GBDT feature frequency} of the sample with the sum of the $k$ decision trees' binary vectors, as:
$$\alpha = \sum_{k} \alpha^{(k)},$$
where $\alpha$ represents the exploitation level of each feature in the GBDT, suggesting the feature preference of the GBDT model on this sample.

\subsubsection*{\textbf{Feature Tokenizer.}} Inspired by the classical language models (e.g., BERT~\cite{kenton2019bert}), recent dominating Transformer-based tabular models~\cite{gorishniy2021revisiting,somepalli2022saint,yan2023t2g} adopted distributed feature representation~\cite{mikolov2013efficient} by embedding tabular values into vector spaces and treating the values as ``unordered'' word vectors. Such Transformer models use \textit{feature tokenizer}~\cite{gorishniy2021revisiting} to process tabular features as follows: Each tabular scalar value is mapped to a vector $e \in \mathbb{R}^d$ with a feature-specific linear projection, where $d$ is the feature hidden dimension. For numerical (continuous) values, the projection weights are multiplied with the value magnitudes. Given $F_1$ numerical features and $F_2$ categorical features, the feature tokenizer outputs feature embedding $E\in \mathbb{R}^{(1+F_1+F_2) \times d}$ by stacking projected features (and an extra $\left [ \text{CLS} \right ] $ token embedding), i.e., $E = \text{stack}\left (\left [ e_{\text{CLS}},e_{\text{num}}^{(1)},\ldots,e_{\text{num}}^{(F_1)},e_{\text{cat}}^{(1)},\ldots,e_{\text{cat}}^{(F_2)}  \right ] \right )$.

\subsection{GBDT Feature Gate} \label{method:gfg}
Early attempts of tabular DNNs tried to emulate behavioral patterns of GBDTs by ensembling neural networks to build \textit{differential tree models}, such as representative models NODE~\cite{popov2019neural} and TabNet~\cite{arik2021tabnet}. However, even realizing decision-tree-like hard feature selection or resorting to complicated Transformer architectures, they were still rapidly submerged in subsequent DNN studies that mainly focused on promotion from deep learning perspectives~\cite{gorishniy2021revisiting,chen2022danets,yan2023t2g}. We seek to rethink this line of work and observe that they achieve hard feature selection with learnable continuous feature masks through DNNs' smooth back-propagation, which may be incompatible with the discrete nature of GBDTs, and hence restrict their potential.

To resolve this issue, we propose \textit{GBDT Feature Gate} (GFG), a GBDT-based feature selector tensorized with GBDT weights to faithfully replicate its feature selection behavior. Specifically, given a GFG initialized by a $T$-tree GBDT, the feature selection process on an $F$-feature sample $x$ ($\hat{E} = \text{GFG}(x) \in \mathbb{R}^{F \times d}$) is formulated as:
\begin{gather}
  E = \text{FeatureTokenizer}(x) \in \mathbb{R}^{F \times d}, \\
 \alpha = \text{GBDTFeatureFrequency}(x) \in \mathbb{R}^{F}, \label{gff}\\
 \hat{\alpha} = \alpha / T \in \mathbb{R}^{F}, \bar{\alpha} = \text{BinarySampler}(\hat{\alpha}) \in \left \{  0, 1\right \}^F, \label{gbdt-freq}\\
 \hat{E}_{:,i} = \begin{cases}
 \bar{\alpha} \odot E_{:,i} \ \ \text{if} \ \textit{training} \\
 \hat{\alpha} \odot E_{:,i} \ \ \text{if} \ \textit{inference}\end{cases},
 i \in \left \{ 1,2,\ldots,d \right \}. \label{feat-gate}
\end{gather}

The extra $\left [ \text{CLS} \right ] $ embedding is omitted in this subsection for notation brevity; in implementation, it is directly concatenated to the head of the gated $\hat{E}$. In Eq.~(\ref{gbdt-freq}), $\hat{\alpha}$ is the normalized GBDT feature frequency that represents the access probabilities of each feature in the $T$-tree GBDT, and $\bar{\alpha}$ is a binary feature mask sampled with the probabilities $\hat{\alpha}$. To incorporate the GBDT's feature preference into the DNN framework, in Eq.~(\ref{feat-gate}), we use sparse feature masks from real GBDT feature access probabilities to perform hard feature selection during training, and use the soft probabilities during inference for deterministic prediction. GFG assists in filtering out unnecessary features according to the GBDT's feature preference, ensuring an oracle selection behavior compared to previous differential tree models in learning feature masks with neural networks.

Since the original GBDT library (we uniformly use XGBoost in this work) has no APIs for efficiently fetching sample-specific GBDT feature frequency in Eq.~(\ref{gff}) and the used backend is incompatible with common DL libraries (e.g., PyTorch), to integrate the GFG module into the parallel DNN framework, we tensorize the behavior of Eq.~(\ref{gff}). Technically, we are inspired by the principle of the Microsoft Hummingbird compiling tools\footnote{https://github.com/microsoft/hummingbird} and extract routing matrices, a series of parameter matrices that contain information of each decision tree's node adjacency and threshold values, from the XGBoost model. Based on the extracted routing matrices, feature access frequency can be simply acquired through alternating tensor multiplication and comparison on input features $x$, and the submodule of Eq.~(\ref{gff}) is initialized with these parameter matrices.

In the actual implementation, we just rapidly train an XGBoost with uniform default hyperparameters provided in~\cite{gorishniy2021revisiting} (regardless of its performance) to initialize and freeze the submodule of Eq.~(\ref{gff}) during the T-MLP initialization step. Other trainable parameters are randomly initialized. Since there are a large number of decision trees to vote the feature preference in a GBDT model, slight hyperparameter modification will not change the overall feature preference trend, and a lightly-trained default XGBoost is always usable enough to guide greedy feature selection. To further speed up the processes in Eqs.~(\ref{gff})-(\ref{gbdt-freq}), we cache the normalized feature frequency $\hat{\alpha}$ for each sample during the first-epoch computation, and reuse the cache in the subsequent model training or inference.

\subsection{Pure MLP Basic Block}
To explore the capability of pure-MLP architecture and keep our tabular model compact, we take inspiration from vision MLPs. We observe that a key factor of their success is the attention-like interaction realized by linear projection and soft gating on features~\cite{tolstikhin2021mlp,guo2022hire,cao2023strip}. Thus, we employ the spatial gating unit (SGU) proposed in~\cite{liu2021pay}, and formulate a simplified pure-MLP block, as:
\begin{gather}
    \hat{E}^{(l+1)} = \text{SGU}(\text{GELU}(\text{LayerNorm}(\hat{E}^{(l)})W_1))W_2 + \hat{E}^{(l)}, \label{basic-block}\\
    \text{SGU}(X) = W_3\text{LayerNorm}(X_{:,:d'}) \odot X_{:,d':} \ . \label{sgu}
\end{gather}

\noindent
The block is similar to a single feed-forward neural network (FFN) in the Transformer with an extra SGU (Eq.~(\ref{sgu})) for feature-level interaction. The main parameters are located in two transformations, i.e., $W_1 \in \mathbb{R}^{d \times 2d'}$ and $W_2 \in \mathbb{R}^{d' \times d}$ in Eq.~(\ref{basic-block}), where $d'$ corresponds to the FFN intermediate dimension size. In Eq.~(\ref{sgu}), $X \in \mathbb{R}^{F \times 2d'}$ denotes the input features of SGU, and $W_3 \in \mathbb{R}^{F \times F}$ is a feature-level transformation to emulate attention operation. Since $d \approx d' \gg F$ in most cases, the model size is determined by $W_1$ and $W_2$, and is comparable to the FFN size. All the bias vectors are omitted for notation brevity.

Analogous to vision data, we treat tabular features and feature embeddings as image pixels and channels. But completely different from vision MLPs, T-MLP is a hybrid framework tailored for economical tabular prediction that performs competitively against tabular Transformers and GBDTs with significantly reduced runtime costs. On most uncomplicated datasets, using only one basic block in T-MLP is enough. In comparison, previous vision MLP studies emphasized architecture engineering and often demanded dozens of blocks in order to be comparable to vision Transformers.

\subsection{Sparsity with User-controllable Pruning}
Inspired by the pre-pruning of GBDTs that controls model complexity and promotes generalization with user-defined hyperparameters (e.g., maximum tree depth, minimum samples per leaf), we design a similar mechanism for T-MLP by leveraging the predominant model compression approach, i.e., DNN \textit{pruning}~\cite{hou2020dynabert,ma2023llm,sun2023simple}, which is widely used in NLP research to trim down over-parameterized language models while maintaining the original reliability~\cite{xia2022structured}.

Specifically, we introduce two fine-grained variables $z_{\text{h}} \in \left \{  0, 1\right \}^d$ and $z_{\text{in}} \in \left \{  0, 1\right \}^{d'}$ to mask parameters from hidden dimension and intermediate dimension, respectively. As the previous FFN pruning in language models~\cite{wang2020structured}, the T-MLP pruning operation can be attained by simply applying the mask variables to the weight matrices, i.e., substituting $W_1$ and $W_2$ with $\text{diag}(z_{\text{h}})W_1$ and $\text{diag}(z_{\text{in}})W_2$ in Eq.~(\ref{basic-block}). We use the classical $l_0$ regularization reparametrized with hard concrete distributions~\cite{louizos2018learning}, and adopt a Lagrangian multiplier objective to achieve the controllable sparsity as in~\cite{xia2022structured}.

Although early attempts of tabular DNNs have considered sparse structures, for example, TabNet~\cite{arik2021tabnet} and NODE~\cite{popov2019neural} built learnable sparse feature masks, and more recently TabCaps~\cite{chen2022tabcaps} and T2G-Former~\cite{yan2023t2g} designed sparse feature interaction, there are two essential differences: (1) existing tabular DNNs only considered sparsity on the feature dimension, while T-MLP introduces sparsity on the input features (Sec.~\ref{method:gfg}) and the hidden dimension (this subsection), which was ignored in previous tabular DNN prediction studies and widely recognized as an over-parameterized facet in NLP practice~\cite{wang2020structured,xia2022structured}; (2) learnable sparsity in existing tabular DNNs is completely coupled and determined by prediction loss functions, while our introduced DNN pruning techniques determine the sparsity based on the user-defined sparsity rate (objective-independent), with the same controllable nature of GBDTs pre-pruning.

In the main experiments (Sec.~\ref{main-exp}), we uniformly fix the target sparsity at 0.33 for T-MLP, i.e., only around 33\% of DNN parameters are retained after training. We further explore the relationship between model sparsity and performance in Sec.~\ref{abl-exp}, and obtain performance boost with suitable parameter pruning, even on T-MLP with one basic block, This implies pervasive over-parameterization in previous tabular DNN designs.

\subsection{Overall Workflow and Efficient Ensemble} \label{ensemble}

The overall T-MLP workflow is as follows: During the training stage, the input tabular features are embedded with the feature tokenizer and discretely selected by the sampled feature mask $\bar{\alpha}$ in Eq.~(\ref{gbdt-freq}); then, they are processed by a single pruned basic block in Eq.~(\ref{basic-block}), and the pruning parameter masks $z_{\text{h}}$ and $z_{\text{in}}$ are sampled with reparameterization on the $l_0$ regularization; the final prediction is made with the $\left [ \text{CLS} \right ]$ token feature using a normal prediction head as in other tabular Transformers, as:
$$\hat{y} = \text{FC}(\text{ReLU}(\text{LayerNorm}(\hat{E}^{(l)}_{\left [ \text{CLS} \right ],:}))),$$
where FC denotes a fully connected layer. We use the cross entropy loss for classification and the mean squared error loss for regression as in previous tabular DNNs. The whole framework is optimized with back-propagation. After training, the parameter masks are directly applied to $W_1$ and $W_2$ by accordingly dropping the pruned hidden and intermediate dimensions. In the inference stage, the input features are softly selected by the normalized GBDT feature frequency $\hat{\alpha}$ in Eq.~(\ref{gbdt-freq}), and processed by the simplified basic block.

Since the T-MLP architecture is compact and computation-friendly with low runtime cost, we further provide an \textit{efficient ensemble} version by simultaneously training three branches with the shared GBDT feature gate from the same initialization point with three fixed learning rates. This produces three different sparse MLPs, inspired by the \textit{model soups} ensemble method~\cite{wortsman2022model}. The final ensemble prediction is the average result of the three branches as in a bagging ensemble model. Since the ensemble learning process can be implemented by simultaneous training and inference with multi-processing programming (e.g., RandomForest~\cite{breiman2001random}), the training duration is not tripled but determined by the slowest converging branch.

\section{Experiments}

In this section, we first compare our T-MLP with advanced DNNs and classical GBDTs on their dominating benchmarks (including 88 datasets for different task types) and analyze from the perspective of cost-effectiveness. Next, we conduct ablation and comparison experiments with multi-facet analysis to evaluate the key designs that make T-MLP effective. Besides, we compare the optimized patterns of common DNNs, GBDTs, and T-MLP by visualizing their decision boundaries to further examine the superiority of T-MLP.

\subsection{Experimental Setup} \label{exp-setup}
\subsubsection*{\textbf{Datasets.}} We use four recent high-quality tabular benchmarks (FT-Transformer\footnote{\url{https://github.com/yandex-research/rtdl-revisiting-models/tree/main}} (FT-T, 11 datasets)~\cite{gorishniy2021revisiting}, T2G-Former\footnote{\url{https://github.com/jyansir/t2g-former/tree/master}} (T2G, 12 datasets)~\cite{yan2023t2g}, SAINT\footnote{\url{https://github.com/somepago/saint/tree/main}} (26 datasets)~\cite{somepalli2022saint}, and Tabular Benchmark\footnote{\url{https://github.com/LeoGrin/tabular-benchmark/tree/main}} (TabBen, 39 datasets)~\cite{grinsztajn2022tree}), considering their elaborated results on extensive baselines and datasets. The FT-T and T2G benchmarks are representative of large-scale tabular datasets, whose sizes vary from 10K to 1,000K and include various DNN baselines. The SAINT benchmark is gathered from the OpenML repository\footnote{\url{https://www.openml.org}}, and is dominated by the pre-trained DNN SAINT, containing balanced task types and diverse GBDTs. TabBen is based on ``typical tabular data'' settings that constrain dataset properties, e.g., the data scale (a maximum data volume of 10K) and the feature number (not high-dimension)~\cite{grinsztajn2022tree}, and the datasets are categorized into several types with combinations of task types and feature characteristics. Notably, on TabBen, GBDTs achieve overwhelming victory, surpassing commonly-used DNNs. Each benchmark represents a specific framework preference. Since several benchmarks have adjusted dataset arrangements in their current repositories (e.g., some datasets were removed and some were added), to faithfully follow and reuse the results, we only retain the datasets reported in the published original papers. We provide detailed benchmark statistical information in Table~\ref{app:bench-info} and discuss benchmark characteristics in Appendix~\ref{benchmark-information}.

\begin{figure}[h]
  \centering
  \includegraphics[width=\linewidth]{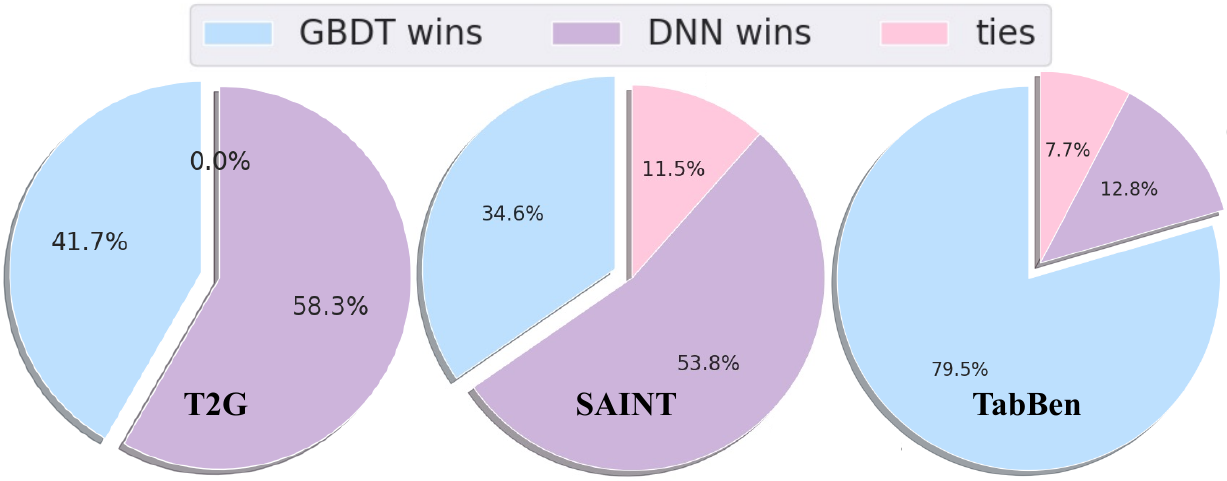}
  \caption{The winning rates of GBDTs and DNNs on three benchmarks, which represent the proportion of each framework achieving the best performance in the benchmarks. It exhibits varying framework preferences among the datasets used in different tabular prediction works.}
  \Description{The illustration of winning rate of GBDTs and DNNs on each benchmark, indicating a varying framework preference of different datasets.}
  \label{win-rate}
\end{figure}

\begin{table*}
\caption{Dataset statistics on four experimental benchmarks. ``\# bin., \# mul., and \# reg.'' are the amounts of binary classification, multi-class classification, and regression datasets. ``\# small, \# middle, \# large, and \# ex. large'' represent the amounts of small ($N \le $ 3K), middle (3K $< N \le$ 10K), large (10K $< N \le$ 100K), and extremely large ($N > $ 100K) datasets, where $N$ denotes the training data size. ``\# wide and \# ex. wide'' are the amounts of wide (32 $< F \le$ 64) and extremely wide ($F >$ 64) datasets, where $F$ is the feature amount. ``bin. metric, mul. metric, and reg. metric'' represent the evaluation metrics used for each task type in the benchmarks. ``R-Squared'' score is the coefficient of determination.} 
\label{app:bench-info}
\centering
\setlength{\tabcolsep}{3.0pt}
\begin{tabular}{@{}l|ccc|cccc|cc|ccc@{}}
\toprule
       & \# bin. & \# mul. & \# reg. & \# small & \# middle & \# large & \# ex. large & \# wide & \# ex. wide & bin. metric & mul. metric & reg. metric \\ \midrule
FT-T~\cite{gorishniy2021revisiting}   & 3      & 4      & 4      & 0       & 0        & 6       & 5           & 2      & 5          & ACC         & ACC         & RMSE                               \\
T2G~\cite{yan2023t2g}    & 3      & 5      & 4      & 0       & 3        & 7       & 2           & 2      & 2          & ACC         & ACC         & RMSE                               \\
SAINT~\cite{somepalli2022saint}  & 9      & 7      & 10     & 10      & 3        & 12      & 1           & 6      & 9          & AUC         & ACC         & RMSE                               \\
TabBen~\cite{grinsztajn2022tree} & 15     & 0      & 24     & 2       & 37       & 0       & 0           & 5      & 2          & ACC         & N/A         & R-Squared                               \\ \bottomrule
\end{tabular}
\end{table*}

\begin{table*}
\caption{Cost-effectiveness comparison on the FT-T benchmark. Classification datasets and regression datasets are evaluated using the accuracy and RMSE metrics, respectively. ``Rank'' denotes the average values (standard deviations) of all the methods across the datasets. ``$T$'' represents the average overhead of the used training time against T-MLP, and ``$T^*$'' compares only the duration before achieving the best validation scores. All the training durations are estimated with the original hyperparameter search settings. ``$P$'' denotes the average parameter number of the best model configuration provided by the FT-T repository. TabNet is not compared considering its different backend (Tensorflow) in the evaluation. The top performances are marked in bold, and the second best ones are underlined (similar marks are used in the subsequent tables).}
\label{tab:main-ftt}
\centering
\setlength{\tabcolsep}{3pt}
\begin{tabular}{@{}l|ccccccccccc|cccc@{}}
\toprule
        & CA $\downarrow$ & AD $\uparrow$ & HE $\uparrow$ & JA $\uparrow$ & HI $\uparrow$ & AL $\uparrow$ & EP $\uparrow$ & YE $\downarrow$ & CO $\uparrow$ & YA $\downarrow$ & MI $\downarrow$ & Rank     & $T$      & $T^*$     & $P$(M)  \\ \midrule
TabNet  & 0.510    & 0.850   & 0.378  & 0.723  & 0.719  & 0.954  & 0.8896 & 8.909    & 0.957  & 0.823    & 0.751    & 9.0 (1.5) & N/A    & N/A    & N/A   \\
SNN     & 0.493    & 0.854  & 0.373  & 0.719  & 0.722  & 0.954  & 0.8975 & 8.895    & 0.961  & 0.761    & 0.751    & 7.8 (1.1) & $\times$42.76  & $\times$24.87  & 1.12  \\
AutoInt & 0.474    & 0.859  & 0.372  & 0.721  & 0.725  & 0.945  & 0.8949 & 8.882    & 0.934  & 0.768    & 0.750    & 7.4 (2.1) & $\times$121.68 & $\times$112.31 & 1.14  \\
GrowNet & 0.487    & 0.857  & N/A    & N/A    & 0.722  & N/A    & 0.8970  & 8.827    & N/A    & 0.765    & 0.751    & N/A      & N/A    & N/A    & N/A   \\
MLP     & 0.499    & 0.852  & 0.383  & 0.719  & 0.723  & 0.954  & 0.8977 & 8.853    & 0.962  & 0.757    & 0.747    & 6.5 (1.7) & $\times$27.41  & $\times$28.46  & 0.55  \\
DCNv2   & 0.484    & 0.853  & 0.385  & 0.716  & 0.723  & 0.955  & 0.8977 & 8.890    & 0.965  & 0.757    & 0.749    & 6.4 (1.8) & $\times$31.15  & $\times$40.65  & 4.17  \\
NODE    & 0.464    & 0.858  & 0.359  & 0.727  & 0.726  & 0.918  & 0.8958 & 8.784    & 0.958  & \textbf{0.753}    & \textbf{0.745}    & 5.4 (3.2) & $\times$386.54 & $\times$353.38 & 16.59 \\
ResNet  & 0.486    & 0.854  & \textbf{0.396}  & \underline{0.728}  & 0.727  & \textbf{0.963}  & 0.8969 & 8.846    & 0.964  & 0.757    & 0.748    & 4.5 (2.2) & $\times$56.20  & $\times$58.46  & 6.16  \\
FT-T    & 0.459    & 0.859  & \underline{0.391}  & \textbf{0.732}  & 0.720   & \underline{0.960}  & \textbf{0.8982} & 8.855    & \textbf{0.970}  & 0.756    & \underline{0.746}    & 3.3 (2.4) & $\times$117.35 & $\times$97.49  & 2.12  \\ \midrule
T-MLP    & \underline{0.447}    & \underline{0.864}  & 0.386  & \underline{0.728}  & \underline{0.729}  & 0.956  & 0.8977 & \underline{8.768}    & 0.968  & 0.756    & 0.747    & \underline{3.1 (0.9)} & $\times$1.00   & $\times$1.00   & 0.79  \\
T-MLP(3) & \textbf{0.438}    & \textbf{0.867}  & 0.386  & \textbf{0.732}  & \textbf{0.730}  & \underline{0.960}  & \underline{0.8978} & \textbf{8.732}    & \underline{0.969}  & \underline{0.755}    & \textbf{0.745}    & \textbf{1.7 (0.8)} & $\times$1.05   & $\times$1.08   & 2.37  \\ \bottomrule
\end{tabular}
\end{table*}

\begin{table*}
\caption{Cost-effectiveness comparison on the T2G benchmark with similar notations as in Table~\ref{tab:main-ftt}. The baseline performances and configurations are also reused from the T2G repository. According to the T2G paper, for the extremely large dataset Year, FT-T and T2G use 50-iteration hyperparameter tuning (HPT), DANet-28 follows its default hyperparameters, and the other baseline results are acquired with 100-iteration HPT.}
\label{tab:main-t2g}
\centering
\setlength{\tabcolsep}{2.5pt}
\begin{tabular}{@{}l|cccccccccccc|cccc@{}}
\toprule
         & GE $\uparrow$ & CH $\uparrow$ & EY $\uparrow$ & CA $\downarrow$ & HO $\downarrow$ & AD $\uparrow$ & OT $\uparrow$ & HE $\uparrow$ & JA $\uparrow$ & HI $\uparrow$ & FB $\downarrow$ & YE $\downarrow$ & Rank & $T$ & $T^*$ & $P$(M)  \\ \midrule
XGBoost  & 0.684 & 0.859 & 0.725 & \textbf{0.436} & 3.169 & \textbf{0.873} & \textbf{0.825} & 0.375 & 0.719 & 0.724 & \textbf{5.359} & 8.850 & 4.3 (3.1)  & $\times$32.78  & $\times$42.88  & N/A   \\
MLP      & 0.586 & 0.858 & 0.611 & 0.499 & 3.173 & 0.854 & 0.810 & 0.384 & 0.720 & 0.720 & 5.943 & 8.849 & 8.3 (1.9)  & $\times$13.73  & $\times$11.45  & 0.64  \\
SNN      & 0.647 & 0.857 & 0.616 & 0.498 & 3.207 & 0.854 & 0.812 & 0.372 & 0.719 & 0.722 & 5.892 & 8.901 & 8.3 (1.5)  & $\times$22.74  & $\times$12.54  & 0.82  \\
TabNet   & 0.600 & 0.850 & 0.621 & 0.513 & 3.252 & 0.848 & 0.791 & 0.379 & 0.723 & 0.720 & 6.559 & 8.916 & 10.2 (2.4) & N/A    & N/A    & N/A   \\
DANet-28 & 0.616 & 0.851 & 0.605 & 0.524 & 3.236 & 0.850 & 0.810 & 0.355 & 0.707 & 0.715 & 6.167 & 8.914 & 10.6 (2.0) & N/A    & N/A    & N/A   \\
NODE     & 0.539 & 0.859 & 0.655 & 0.463 & 3.216 & 0.858 & 0.804 & 0.353 & 0.728 & 0.725 & 5.698 & 8.777 & 7.0 (3.0)  & $\times$329.79 & $\times$288.21 & 16.95 \\
AutoInt  & 0.583 & 0.855 & 0.611 & 0.472 & 3.147 & 0.857 & 0.801 & 0.373 & 0.721 & 0.725 & 5.852 & 8.862 & 8.1 (2.0)  & $\times$68.30  & $\times$55.52  & 0.06  \\
DCNv2    & 0.557 & 0.857 & 0.614 & 0.489 & 3.172 & 0.855 & 0.802 & \underline{0.386} & 0.716 & 0.722 & 5.847 & 8.882 & 8.4 (2.0)  & $\times$24.40  & $\times$21.63  & 2.30  \\
FT-T     & 0.613 & 0.861 & 0.708 & 0.460 & \underline{3.124} & 0.857 & 0.813 & \textbf{0.391} & \underline{0.732} & \underline{0.731} & 6.079 & 8.852 & 4.7 (2.6)  & $\times$64.68  & $\times$50.90  & 2.22  \\
T2G      & 0.656 & \underline{0.863} & \textbf{0.782} & 0.455 & 3.138 & 0.860 & 0.819 & \textbf{0.391} & \textbf{0.737} & \textbf{0.734} & 5.701 & 8.851 & \underline{3.1 (1.7)}  & $\times$88.93  & $\times$87.04  & 1.19  \\ \midrule
T-MLP     & \underline{0.706} & 0.862 & 0.717 & 0.449 & 3.125 & 0.864 & 0.814 & \underline{0.386} & 0.728 & 0.729 & 5.667 & \underline{8.768} & 3.3 (0.9)  & $\times$1.00   & $\times$1.00   & 0.72  \\
T-MLP(3)  & \textbf{0.714} & \textbf{0.866} & \underline{0.747} & \underline{0.438} & \textbf{3.063} & \underline{0.867} & \underline{0.823} & \underline{0.386} & \underline{0.732} & 0.730 & \underline{5.629} & \textbf{8.732} & \textbf{1.9 (0.8)}  & $\times$1.09   & $\times$1.11   & 2.16  \\ \bottomrule
\end{tabular}
\end{table*}

\subsubsection*{\textbf{Implementation Details.}} 
We implement our T-MLP model and Python package using PyTorch on Python 3.10. Since the reported baseline training durations on the original benchmarks are estimated under different runtime environments and using different evaluation codes, and do not consider hyperparameter tuning (HPT) budgets, for uniform comparison of training costs, we encapsulate the experimental baselines with the same sklearn-style APIs as T-MLP in our built package, and conduct all the experiments on NVIDIA A100 PCIe 40GB. All the hyperparameter spaces and iteration numbers of the baselines follow the settings in the original papers to emulate the tuning process of each baseline. For T-MLP, we use fixed hyperparameters as the model is trained only once. The XGBoost used for T-MLP's GBDT Feature Gate is in default configuration as in~\cite{gorishniy2021revisiting}. In experiments, each single T-MLP uses one basic block for most datasets if without special specification. We uniformly use a learning rate of 1e-4 for a single T-MLP and learning rates of 1e-4, 5e-4, and 1e-3 for the three branches in the T-MLP ensemble (group ``T-MLP(3)''). We reuse the same data splits as in the original benchmarks. The baseline performances are inherited from the reported benchmark results, and the baseline capacities are calculated based on the best model configurations provided in the corresponding paper repositories. Detailed information of the runtime environment and hyperparameters is given in Appendix~\ref{hyper-info}.

\subsubsection*{\textbf{Compared Methods.}} 
On the four benchmarks, we compare our T-MLP (the single-model and 3-model-ensemble versions) with: (1) known non-pre-trained DNNs: MLP, ResNet, SNN~\cite{klambauer2017self}, GrowNet~\cite{badirli2020gradient}, TabNet~\cite{arik2021tabnet}, NODE~\cite{popov2019neural}, AutoInt~\cite{song2019autoint}, DCNv2~\cite{wang2021dcn}, TabTransformer~\cite{huang2020tabtransformer}, DANets~\cite{chen2022danets}, FT-Transformer (FT-T)~\cite{gorishniy2021revisiting}, and T2G-Former (T2G)~\cite{yan2023t2g}; (2) pre-trained DNN: SAINT~\cite{somepalli2022saint}; (3) GBDT models: XGBoost~\cite{chen2016xgboost}, CatBoost~\cite{prokhorenkova2018catboost}, LightGBM~\cite{ke2017lightgbm}, GradientBoostingTree (GBT), HistGradientBoostingTree (HistGBT), and other traditional non-deep machine learning methods like RandomForest (RF)~\cite{breiman2001random}. For other unmentioned baselines, please refer to Appendix~\ref{baseline-info}. In the experiment tables below, ``T-MLP'' denotes a single T-MLP and ``T-MLP(3)'' denotes the ensemble version with three branches.

\subsection{Main Results and Analysis} \label{main-exp}

In Table~\ref{tab:main-ftt} to Table~\ref{tab:main-whytree}, the baseline results are based on heavy HPT, and are obtained from respectively reported benchmarks. All the T-MLP results are based on default hyperparameters.

\subsubsection*{\textbf{Comparison with Advanced DNNs.}} 
Tables~\ref{tab:main-ftt} and~\ref{tab:main-t2g} report detailed performances and runtime costs on the FT-T and T2G benchmarks for comparison of our T-MLP versions and bespoke tabular DNNs~\cite{gorishniy2021revisiting,yan2023t2g}. The baseline results in these tables are based on 50 (for complicated models on large datasets, e.g., FT-Transformer on the Year dataset) or 100 (the other cases) iterations of HPT except special models (default NODE for the datasets with large class numbers and default DANets for all datasets). An overall trend that one may observe is that the single T-MLP is able to achieve competitive results as the state-of-the-art DNNs on each benchmark, and a simple ensemble of three T-MLPs (i.e., ``T-MLP(3)'') exhibits even better performances with significantly reduced training costs. Specifically, benefiting from fixed hyperparameters and simple structures, the single T-MLP achieves obvious speedup and reduces training durations by orders of magnitude compared to the powerful DNNs, and is also more training-friendly than XGBoost, a representative GBDT that highly relies on heavy HPT. Besides, we observe only about 10\% training duration increase in T-MLP ensemble since we adopt multiprocessing programming to simultaneously train the three T-MLPs (see Sec.~\ref{ensemble}) and thus the training time depends on the slowest converging sub-model. In the implementation details (Sec.~\ref{exp-setup}), the single T-MLP uses the smallest learning rate in the three sub-models, and hence the convergence time of T-MLP ensemble often approximates that of the single T-MLP. From the perspective of model storage, as expected, the size of the single T-MLP is comparable to the average level of naive MLPs across the datasets and its size variation is stable (see Table~\ref{tab:cost-perform}), since the block number, hidden dimension size, and sparsity rate are all fixed. In Sec.~\ref{abl-exp}, we will further analyze the impact of model sparsity and theoretical complexity of the model parameters. 

\renewcommand{\arraystretch}{0.8}
\begin{table}
\caption{The average values (standard deviations) of all the method ranks on the SAINT benchmark of three task types. $|D|$ is the dataset number in each group. Notably, all the baseline results are based on HPT, and SAINT variants need further training budgets on pre-training and data augmentation. More detailed results are given in the Appendix.}
\label{tab:main-saint}
\centering
\setlength{\tabcolsep}{2pt}
\begin{tabular}{@{}l|c|c|c@{}}
\toprule
Task Type: & \begin{tabular}[c]{@{}c@{}}Binclass\\ ($|D|$=9)\end{tabular} & \begin{tabular}[c]{@{}c@{}}Multiclass\\ ($|D|$=7)\end{tabular} & \begin{tabular}[c]{@{}c@{}}Regression\\ ($|D|$=10)\end{tabular} \\ \midrule
RF         & 7.8 (3.3)          & 7.3 (2.2)          & 9.1 (4.2)          \\
ExtraTrees & 7.8 (3.8)          & 9.6 (1.9)          & 8.6 (3.5)          \\
KNeighborsDist    & 13.7 (0.7)         & 11.6 (3.5)         & 12.9 (1.8)         \\
KNeighborsUnif    & 14.4 (0.5)         & 12.4 (3.4)         & 14.0 (1.0)         \\
LightGBM   & 5.7 (3.3)          & \underline{3.9 (2.8)}    & 6.5 (3.2)          \\
XGBoost    & \underline{4.2 (2.8)}    & 6.7 (3.5)          & 7.3 (2.9)          \\
CatBoost   & \textbf{3.9 (2.8)} & 7.2 (2.4)          & 5.6 (2.7)          \\
MLP        & 10.7 (1.8)         & 10.1 (3.9)         & N/A               \\
NeuralNetFastAI   & N/A               & N/A               & 11.9 (2.2)         \\
TabNet     & 13.2 (2.0)         & 13.5 (1.1)         & 10.2 (4.5)         \\
TabTransformer      & 10.8 (1.4)         & 10.0 (3.6)         & 10.0 (2.9)         \\
SAINT-s    & 7.8 (2.4)          & 7.9 (6.1)          & 4.7 (3.8)          \\
SAINT-i    & 7.2 (2.6)          & 7.1 (2.7)          & 5.9 (3.5)          \\
SAINT      & \underline{4.2 (2.7)}    & 5.2 (2.2)          & \textbf{4.2 (2.3)} \\ \midrule
T-MLP       & 4.6 (2.8)          & 4.6 (3.0)          & \underline{4.6 (3.3)}    \\
T-MLP(3)    & \textbf{3.9 (1.9)} & \textbf{2.9 (2.5)} & 5.0 (2.9)          \\ \bottomrule
\end{tabular}
\end{table}
\renewcommand{\arraystretch}{1}

\subsubsection*{\textbf{Comparison with Pre-trained DNNs.}} 
Table~\ref{tab:main-saint} reports the means and standard deviations of model ranks on the SAINT benchmark~\cite{somepalli2022saint}. Surprisingly, we find that the simple pure MLP-based T-MLP outperforms Transformer-based SAINT variants (SAINT-s and SAINT-i) and is comparable with SAINT on all the three task types. It is worth noting that SAINT and its variants adopt complicated inter-sample attention and self-supervised pre-training along with HPT on parameters of the training process. Moreover, T-MLP ensemble even achieves stable results that are competitive to tuned GBDTs (i.e., XGBoost, CatBoost, LightGBM) and surpasses the pre-trained SAINT on classification tasks. Since the detailed HPT conditions (i.e., iteration times, HPT methods, parameter sampling distributions) are not reported, we do not estimate specific training costs.

\subsubsection*{\textbf{Comparison with Extensively Tuned GBDTs.}} 
Table~\ref{tab:main-whytree} compares T-MLP on the typically GPDTs-dominating benchmark TabBen~\cite{grinsztajn2022tree}, on which GBDT frameworks completely outperform various DNNs across all types of datasets. Results of each baseline on TabBen are obtained with around 400 iterations of heavy HPT, almost representing the ultimate performances with unlimited computation resources and budgets. As expected, when extensively tuned XGBoost is available, the single T-MLP is eclipsed, but it is still competitive to the other ensemble tree models (i.e., RF, GBT, HistGBT) and superior to the compared DNNs. Further, we find that T-MLP ensemble is able to be comparable to the ultimate XGBoost in all the four dataset types with similar rank stability, serving as a candidate for a tuned XGBoost alternative. More significantly, in the experiments, each T-MLP (or T-MLP ensemble) employs a tensorized XGBoost trained in default configuration (see implementation details in Sec.~\ref{exp-setup}), and all the other hyperparameters are fixed; thus T-MLP and its ensemble have potential capability of a higher performance ceiling by HPT or selecting other GBDTs as the feature gate.

In summary, we empirically show the strong potential of our hybrid framework to achieve flexible and generalized data adaptability with various tabular preferences (tabular data preferring advanced DNNs, pre-training, or GBDTs).
Based on the impressive economical performance-cost trade-off and friendly training process, T-MLP can serve as a promising tabular model framework for real-world applications, especially under limited computation budgets.

\renewcommand{\arraystretch}{0.8}
\begin{table}
\caption{The average values (standard deviations) of all the method ranks on TabBen (four dataset types). ``Num.'' and ``Cat.'' denote numerical datasets (all features are numerical) and categorical datasets (some features are categorical), respectively. ``Classif.'' and ``Reg.'' denote classification and regression tasks. ``Num. Reg.'' group includes only results of regression on numerical datasets (similar notations are for the others). $|D|$ is the dataset number in each group. Baseline test results are obtained based on the best validation results during $\sim$400 iterations of HPT (according to the TabBen paper and repository). Detailed results are given in the Appendix.}
\label{tab:main-whytree}
\centering
\setlength{\tabcolsep}{2pt}
\begin{tabular}{@{}l|c|c|c|c@{}}
\toprule
Dataset Type:               & \begin{tabular}[c]{@{}c@{}}Num. Classif. \\ ($|D|$=9)\end{tabular} & \begin{tabular}[c]{@{}c@{}}Num. Reg. \\ ($|D|$=14)\end{tabular} & \begin{tabular}[c]{@{}c@{}}Cat. Classif. \\ ($|D|$=6)\end{tabular} & \begin{tabular}[c]{@{}c@{}}Cat. Reg. \\ ($|D|$=10)\end{tabular} \\ \midrule
MLP                      & 8.4 (0.8)            & N/A               & N/A                 & N/A               \\
ResNet                   & 6.9 (1.9)            & 6.5 (1.9)          & 7.8 (1.0)            & 7.7 (0.5)          \\
FT-T                     & 5.7 (1.9)            & 5.5 (2.3)          & 5.5 (2.2)            & 6.7 (1.1)          \\
SAINT                    & 6.9 (1.4)            & 5.5 (2.2)          & 8.0 (1.1)            & N/A               \\
GBT     & 4.7 (2.0)            & 4.3 (1.7)          & 5.2 (2.3)            & 4.3 (1.1)          \\
HistGBT & N/A                 & N/A               & 5.2 (2.3)            & 4.3 (1.3)          \\
RF             & 4.6 (2.1)            & 4.8 (2.2)          & 4.0 (3.2)            & 5.8 (1.9)          \\
XGBoost                  & \underline{2.6 (1.4)}      & \textbf{2.4 (1.5)} & \textbf{2.8 (1.5)}   & \underline{2.1 (1.0)}    \\ \midrule
T-MLP                     & 3.2 (1.6)            & 4.3 (1.9)          & 3.5 (2.3)            & 3.6 (1.4)          \\
T-MLP(3)                  & \textbf{2.1 (1.4)}   & \underline{2.7 (1.5)}    & \underline{3.0 (1.3)}      & \textbf{1.8 (0.7)} \\ \bottomrule
\end{tabular}
\end{table}
\renewcommand{\arraystretch}{1}

\subsection{What Makes T-MLP Cost-effective?} \label{abl-exp}

Table~\ref{tab:main-abl} reports ablation and comparison experimental results of T-MLP on several classification and regression datasets (i.e., California Housing (CA)~\cite{pace1997sparse}, Adult (AD)~\cite{kohavi1996scaling}, Higgs (HI)~\cite{baldi2014searching}, and Year (YE)~\cite{bertin2011million}) in various data scales (given in parentheses).

\renewcommand{\arraystretch}{0.9}
\begin{table}
\caption{Main ablation and comparison on classical tables in various task types and data scales. The top 4 rows: ablations on key designs in the T-MLP framework. The bottom 2 rows: results of T-MLP with neural network feature gate (NN FG).}
\label{tab:main-abl}
\centering
\setlength{\tabcolsep}{2pt}
\begin{tabular}{@{}l|cccc@{}}
\toprule
Dataset:         & CA (21K) $\downarrow$ & AD (49K) $\uparrow$ & HI (98K) $\uparrow$ & YE (515K) $\downarrow$ \\ \midrule
T-MLP             & 0.4471  & 0.864   & 0.729   & 8.768    \\
w/o sparsity     & 0.4503  & 0.857   & 0.726   & 8.887    \\
w/o GBDT FG     & 0.4539  & 0.859   & 0.728   & 8.799    \\
w/o both         & 0.4602  & 0.856   & 0.724   & 8.896    \\ \midrule
T-MLP (NN FG)     & 0.4559  & 0.852   & 0.718   & 8.925    \\
w/o sparsity     & 0.4557  & 0.840   & 0.713   & 8.936    \\ \bottomrule
\end{tabular}
\end{table}
\renewcommand{\arraystretch}{1}

\subsubsection*{\textbf{Main Ablations.}} 
The top four rows in Table~\ref{tab:main-abl} report the impact of two key designs in a single T-MLP. An overall observation is that both the structure sparsity and GBDT feature gate (FG) contribute to performance enhancement of T-MLP. From the perspective of data processing, GBDT FG brings local sparsity through sample-specific feature selection, and the sparse MLP structure offers global sparsity shared by all samples. Interestingly, we find that the impact of GBDT FG is more profound on the CA dataset. A possible explanation is that the feature amount of CA (8 features) is relatively small compared to the others (14, 28, and 90 features in AD, HI, and YE, respectively) and the average feature importance may be relatively large; thus, the CA results are more likely to be affected by feature selection. For the datasets with larger feature amounts, selecting effective features is likely to be more difficult.

\subsubsection*{\textbf{Greedy Feature Selection.}} 
We notice a recent attempt on sample-specific sparsity for biomedical tables using a gating network; it was originally designed for low-sample-size tabular settings and helped prediction interpretability in the biomedical domain~\cite{yang2022locally}. We use its code and build a T-MLP version by substituting GBDT FG with the neural network feature gate (NN FG) for comparison. The bottom two rows of Table~\ref{tab:main-abl} report the results. As expected, on the smallest dataset CA, NN FG can boost performance by learning to select informative features, but such a feature gating strategy consistently hurts the performance as data scales increase. This may be due to (1) large datasets demand more complicated structures to learn the meticulous feature selection, (2) the discrete nature of the selection behavior is incompatible with smooth optimization patterns of neural networks, and (3) DNNs' confirmation bias~\cite{tarvainen2017mean} may mislead the learning process, i.e., NN FG will be ill-informed once the subsequent neural network captures wrong patterns. In contrast, GBDT FG always selects features greedily as real GBDTs, which is conservative and generally reasonable. Besides, the complicated sub-tree structures are more complete for the selection action. 

\subsubsection*{\textbf{Sparsity Promotes Tabular DNNs.}} 
\begin{figure}[t]
  \centering
  \includegraphics[width=\linewidth]{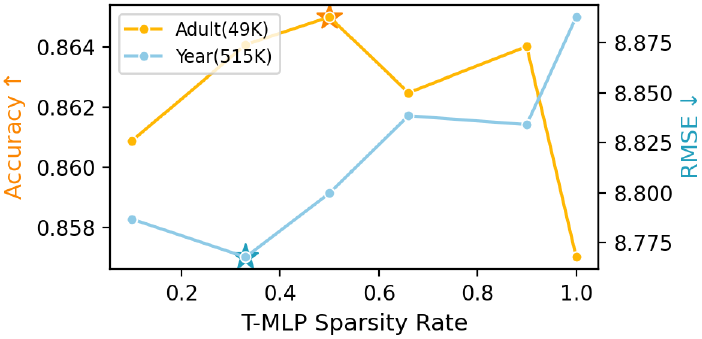}
  \caption{Performance variation plots on the Adult and Year datasets with respect to variations of T-MLP sparsity. All the best results are achieved with suitable sparsity.}
  \Description{Result variations on classification and regression datasets with respect sparsity rates of T-MLP show that, for tabular DNNs, it is common to be over-parameterized and suitable model pruning can promote performances.}
  \label{sparsity-exp}
\end{figure}

Fig.~\ref{sparsity-exp} plots performance variations on two classification/regression tasks with respect to T-MLP sparsity. Different from the pruning techniques in NLP that aim to trim down model sizes while maintaining the ability of the original models, we find that suitable model sparsity often promotes tabular prediction, but both excessive and insufficient sparsity cannot achieve the best results. The results empirically indicate that, compared to DNN pruning in large pre-trained models for unstructured data, in the tabular data domain, the pruning has the capability to promote non-large tabular DNNs as GBDTs' beneficial sparse structures achieved by tree pre-pruning, and the hidden dimension in tabular DNNs is commonly over-parameterized.

\subsection{Superiority Interpretability of T-MLP} \label{superiority-interpret}

\begin{figure}[h]
  \centering
  \includegraphics[width=\linewidth]{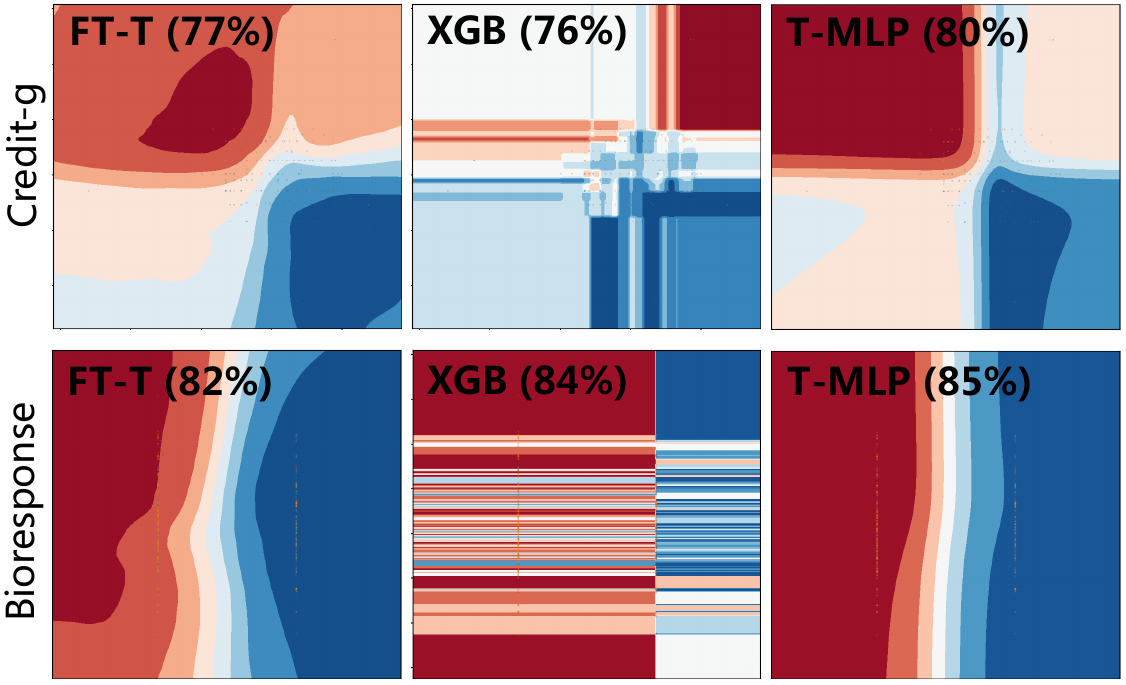}
  \caption{Decision boundary visualization of FT-Transformer (FT-T), XGBoost, and a single-block T-MLP on the Bioresponse and Credit-g datasets, using two most important features. Different colors represent distinct categories, while the varying shades of colors indicate the predicted probabilities.}
  \Description{The decision boundaries of common DNNs, GBDTs and single T-MLP exhibit distinct data patterns captured by the three frameworks, it can be clearly seen T-MLP absorbs both pattern characteristics from DNNs and GBDTs, giving an intermediate state, which is caused by hybrid optimization that combines smooth back-propagation with greedy feature selection and sparse neural architecture space.}
  \label{interpretability}
\end{figure}

In Fig.~\ref{interpretability}, we visualize decision boundaries of FT-Transformer, XGBoost, and the single T-MLP to inspect data patterns captured by these three methods. The two most important features are selected by mutual information (estimated with the Scikit Learn package). Different from common DNNs and GBDTs, T-MLP exhibits a novel intermediate pattern that combines characteristics from both DNNs and GBDTs. Compared to DNNs, T-MLP yields grid-like boundaries whose edges are often orthogonal to feature surfaces as GBDTs, and the complexity is essentially simplified with pruned sparse architectures. Besides, T-MLP is able to capture tree-model-like sub-patterns (see T-MLP on Credit-g), while DNNs manage only main patterns. Hence, DNNs are overfit-sensitive due to their relatively irregular boundaries and neglecting of fine-grained sub-patterns. Compared to GBDTs with jagged boundaries and excessively split sub-patterns, T-MLP holds very smooth vertices at the intersection of boundaries (see T-MLP on Credit-g). Notably, T-MLP can decide conditional split points like GBDT feature splitting (orthogonal edges at feature surfaces) through a smooth process (see T-MLP boundary edges on Bioresponse, from top to bottom, in which the split point on the horizontal feature is conditionally changed with respect to the vertical feature in a smooth manner, while XGBoost is hard to attain such dynamical split points). Overall, T-MLP possesses both the advantages to be overfit-resistant, which helps provide its superiority on both GBDT- and DNN-favored datasets.

\section{Conclusions}
In this paper, we proposed T-MLP, a novel hybrid framework attaining the advantages of both GBDTs and DNNs to address the model selection dilemma in tabular prediction tasks. We combined a tensorized GBDT feature gate, DNN pruning techniques, and a vanilla back-propagation optimizer to develop a simple yet efficient and widely effective MLP model. Experiments on diverse benchmarks showed that, with significantly reduced runtime costs, T-MLP has the generalized adaptability to achieve considerably competitive results regardless of dataset-specific framework preferences. We expect that our T-MLP will serve as a practical method for economical tabular prediction as well as in broad applications, and help advance research on hybrid tabular models.

\begin{acks}
This research was partially supported by National Natural Science Foundation of China under grants No. 62176231, Zhejiang Key R\&D Program of China under grant No. 2023C03053 and No. 2024SSYS0026.
\end{acks}

\balance
\bibliographystyle{ACM-Reference-Format}
\bibliography{sample-base}


\begin{thebibliography}{69}

%%% the Web address.

\ifx \showCODEN    \undefined \def \showCODEN     #1{\unskip}     \fi
\ifx \showDOI      \undefined \def \showDOI       #1{#1}\fi
\ifx \showISBNx    \undefined \def \showISBNx     #1{\unskip}     \fi
\ifx \showISBNxiii \undefined \def \showISBNxiii  #1{\unskip}     \fi
\ifx \showISSN     \undefined \def \showISSN      #1{\unskip}     \fi
\ifx \showLCCN     \undefined \def \showLCCN      #1{\unskip}     \fi
\ifx \shownote     \undefined \def \shownote      #1{#1}          \fi
\ifx \showarticletitle \undefined \def \showarticletitle #1{#1}   \fi
\ifx \showURL      \undefined \def \showURL       {\relax}        \fi
% invisible to TeX
\providecommand\bibfield[2]{#2}
\providecommand\bibinfo[2]{#2}
\providecommand\natexlab[1]{#1}
\providecommand\showeprint[2][]{arXiv:#2}

\bibitem[Altman(1992)]%
        {altman1992introduction}
\bibfield{author}{\bibinfo{person}{Naomi~S Altman}.} \bibinfo{year}{1992}\natexlab{}.
\newblock \showarticletitle{An introduction to kernel and nearest-neighbor nonparametric regression}.
\newblock \bibinfo{journal}{\emph{The American Statistician}} \bibinfo{volume}{46}, \bibinfo{number}{3} (\bibinfo{year}{1992}), \bibinfo{pages}{175--185}.
\newblock


\bibitem[Arik and Pfister(2021)]%
        {arik2021tabnet}
\bibfield{author}{\bibinfo{person}{Sercan~{\"O} Arik} {and} \bibinfo{person}{Tomas Pfister}.} \bibinfo{year}{2021}\natexlab{}.
\newblock \showarticletitle{{TabNet}: Attentive interpretable tabular learning}. In \bibinfo{booktitle}{\emph{AAAI}}. \bibinfo{pages}{6679--6687}.
\newblock


\bibitem[Aziz et~al\mbox{.}(2022)]%
        {aziz2022machine}
\bibfield{author}{\bibinfo{person}{Saqib Aziz}, \bibinfo{person}{Michael Dowling}, \bibinfo{person}{Helmi Hammami}, {and} \bibinfo{person}{Anke Piepenbrink}.} \bibinfo{year}{2022}\natexlab{}.
\newblock \showarticletitle{Machine learning in finance: A topic modeling approach}.
\newblock \bibinfo{journal}{\emph{European Financial Management}} (\bibinfo{year}{2022}).
\newblock


\bibitem[Badirli et~al\mbox{.}(2020)]%
        {badirli2020gradient}
\bibfield{author}{\bibinfo{person}{Sarkhan Badirli}, \bibinfo{person}{Xuanqing Liu}, \bibinfo{person}{Zhengming Xing}, \bibinfo{person}{Avradeep Bhowmik}, \bibinfo{person}{Khoa Doan}, {and} \bibinfo{person}{Sathiya~S Keerthi}.} \bibinfo{year}{2020}\natexlab{}.
\newblock \showarticletitle{Gradient boosting neural networks: {GrowNet}}.
\newblock \bibinfo{journal}{\emph{arXiv preprint arXiv:2002.07971}} (\bibinfo{year}{2020}).
\newblock


\bibitem[Baldi et~al\mbox{.}(2014)]%
        {baldi2014searching}
\bibfield{author}{\bibinfo{person}{Pierre Baldi}, \bibinfo{person}{Peter Sadowski}, {et~al\mbox{.}}} \bibinfo{year}{2014}\natexlab{}.
\newblock \showarticletitle{Searching for exotic particles in high-energy physics with deep learning}.
\newblock \bibinfo{journal}{\emph{Nature Communications}} \bibinfo{volume}{5}, \bibinfo{number}{1} (\bibinfo{year}{2014}), \bibinfo{pages}{4308}.
\newblock


\bibitem[Bertin-Mahieux et~al\mbox{.}(2011)]%
        {bertin2011million}
\bibfield{author}{\bibinfo{person}{Thierry Bertin-Mahieux}, \bibinfo{person}{Daniel~PW Ellis}, \bibinfo{person}{Brian Whitman}, {and} \bibinfo{person}{Paul Lamere}.} \bibinfo{year}{2011}\natexlab{}.
\newblock \showarticletitle{The million song dataset}. In \bibinfo{booktitle}{\emph{ISMIR}}.
\newblock


\bibitem[Borisov et~al\mbox{.}(2022a)]%
        {borisov2022deep}
\bibfield{author}{\bibinfo{person}{Vadim Borisov}, \bibinfo{person}{Tobias Leemann}, \bibinfo{person}{Kathrin Se{\ss}ler}, \bibinfo{person}{Johannes Haug}, \bibinfo{person}{Martin Pawelczyk}, {and} \bibinfo{person}{Gjergji Kasneci}.} \bibinfo{year}{2022}\natexlab{a}.
\newblock \showarticletitle{Deep neural networks and tabular data: A survey}.
\newblock \bibinfo{journal}{\emph{IEEE Transactions on Neural Networks and Learning Systems}} (\bibinfo{year}{2022}).
\newblock


\bibitem[Borisov et~al\mbox{.}(2022b)]%
        {borisov2022language}
\bibfield{author}{\bibinfo{person}{Vadim Borisov}, \bibinfo{person}{Kathrin Sessler}, \bibinfo{person}{Tobias Leemann}, \bibinfo{person}{Martin Pawelczyk}, {and} \bibinfo{person}{Gjergji Kasneci}.} \bibinfo{year}{2022}\natexlab{b}.
\newblock \showarticletitle{Language Models are Realistic Tabular Data Generators}. In \bibinfo{booktitle}{\emph{ICLR}}.
\newblock


\bibitem[Breiman(2001)]%
        {breiman2001random}
\bibfield{author}{\bibinfo{person}{Leo Breiman}.} \bibinfo{year}{2001}\natexlab{}.
\newblock \showarticletitle{Random forests}.
\newblock \bibinfo{journal}{\emph{Machine learning}}  \bibinfo{volume}{45} (\bibinfo{year}{2001}), \bibinfo{pages}{5--32}.
\newblock


\bibitem[Brown et~al\mbox{.}(2020)]%
        {brown2020language}
\bibfield{author}{\bibinfo{person}{Tom Brown}, \bibinfo{person}{Benjamin Mann}, \bibinfo{person}{Nick Ryder}, \bibinfo{person}{Melanie Subbiah}, \bibinfo{person}{Jared~D Kaplan}, \bibinfo{person}{Prafulla Dhariwal}, \bibinfo{person}{Arvind Neelakantan}, \bibinfo{person}{Pranav Shyam}, \bibinfo{person}{Girish Sastry}, \bibinfo{person}{Amanda Askell}, {et~al\mbox{.}}} \bibinfo{year}{2020}\natexlab{}.
\newblock \showarticletitle{Language models are few-shot learners}. In \bibinfo{booktitle}{\emph{NeurIPS}}, Vol.~\bibinfo{volume}{33}. \bibinfo{pages}{1877--1901}.
\newblock


\bibitem[Cao et~al\mbox{.}(2023)]%
        {cao2023strip}
\bibfield{author}{\bibinfo{person}{Guiping Cao}, \bibinfo{person}{Shengda Luo}, \bibinfo{person}{Wenjian Huang}, \bibinfo{person}{Xiangyuan Lan}, \bibinfo{person}{Dongmei Jiang}, \bibinfo{person}{Yaowei Wang}, {and} \bibinfo{person}{Jianguo Zhang}.} \bibinfo{year}{2023}\natexlab{}.
\newblock \showarticletitle{{Strip-MLP}: Efficient Token Interaction for Vision {MLP}}. In \bibinfo{booktitle}{\emph{ICCV}}. \bibinfo{pages}{1494--1504}.
\newblock


\bibitem[Chen et~al\mbox{.}(2022a)]%
        {chen2022tabcaps}
\bibfield{author}{\bibinfo{person}{Jintai Chen}, \bibinfo{person}{KuanLun Liao}, \bibinfo{person}{Yanwen Fang}, \bibinfo{person}{Danny Chen}, {and} \bibinfo{person}{Jian Wu}.} \bibinfo{year}{2022}\natexlab{a}.
\newblock \showarticletitle{{TabCaps}: A Capsule Neural Network for Tabular Data Classification with {BoW} Routing}. In \bibinfo{booktitle}{\emph{ICLR}}.
\newblock


\bibitem[Chen et~al\mbox{.}(2022b)]%
        {chen2022danets}
\bibfield{author}{\bibinfo{person}{Jintai Chen}, \bibinfo{person}{Kuanlun Liao}, \bibinfo{person}{Yao Wan}, \bibinfo{person}{Danny~Z Chen}, {and} \bibinfo{person}{Jian Wu}.} \bibinfo{year}{2022}\natexlab{b}.
\newblock \showarticletitle{{DANets}: Deep abstract networks for tabular data classification and regression}. In \bibinfo{booktitle}{\emph{AAAI}}.
\newblock


\bibitem[Chen et~al\mbox{.}(2023b)]%
        {chen2023excelformer}
\bibfield{author}{\bibinfo{person}{Jintai Chen}, \bibinfo{person}{Jiahuan Yan}, \bibinfo{person}{Danny~Ziyi Chen}, {and} \bibinfo{person}{Jian Wu}.} \bibinfo{year}{2023}\natexlab{b}.
\newblock \showarticletitle{{ExcelFormer}: A Neural Network Surpassing {GBDTs} on Tabular Data}.
\newblock \bibinfo{journal}{\emph{arXiv preprint arXiv:2301.02819}} (\bibinfo{year}{2023}).
\newblock


\bibitem[Chen et~al\mbox{.}(2023a)]%
        {chen2023tsmixer}
\bibfield{author}{\bibinfo{person}{Si-An Chen}, \bibinfo{person}{Chun-Liang Li}, \bibinfo{person}{Nate Yoder}, \bibinfo{person}{Sercan~O Arik}, {and} \bibinfo{person}{Tomas Pfister}.} \bibinfo{year}{2023}\natexlab{a}.
\newblock \showarticletitle{{TSMixer}: An All-{MLP} architecture for time series forecasting}.
\newblock \bibinfo{journal}{\emph{arXiv preprint arXiv:2303.06053}} (\bibinfo{year}{2023}).
\newblock


\bibitem[Chen and Guestrin(2016)]%
        {chen2016xgboost}
\bibfield{author}{\bibinfo{person}{Tianqi Chen} {and} \bibinfo{person}{Carlos Guestrin}.} \bibinfo{year}{2016}\natexlab{}.
\newblock \showarticletitle{{XGBoost}: A scalable tree boosting system}. In \bibinfo{booktitle}{\emph{Proceedings of the 22nd ACM SIGKDD International Conference on Knowledge Discovery and Data Mining}}. \bibinfo{pages}{785--794}.
\newblock


\bibitem[Covington et~al\mbox{.}(2016)]%
        {covington2016deep}
\bibfield{author}{\bibinfo{person}{Paul Covington}, \bibinfo{person}{Jay Adams}, {and} \bibinfo{person}{Emre Sargin}.} \bibinfo{year}{2016}\natexlab{}.
\newblock \showarticletitle{Deep neural networks for {YouTube} recommendations}. In \bibinfo{booktitle}{\emph{Proceedings of the 10th ACM Conference on Recommender Systems}}.
\newblock


\bibitem[Friedman(2001)]%
        {friedman2001greedy}
\bibfield{author}{\bibinfo{person}{Jerome~H Friedman}.} \bibinfo{year}{2001}\natexlab{}.
\newblock \showarticletitle{Greedy function approximation: A gradient boosting machine}.
\newblock \bibinfo{journal}{\emph{Annals of Statistics}} (\bibinfo{year}{2001}).
\newblock


\bibitem[Fusco et~al\mbox{.}(2023)]%
        {fusco-etal-2023-pnlp}
\bibfield{author}{\bibinfo{person}{Francesco Fusco}, \bibinfo{person}{Damian Pascual}, \bibinfo{person}{Peter Staar}, {and} \bibinfo{person}{Diego Antognini}.} \bibinfo{year}{2023}\natexlab{}.
\newblock \showarticletitle{{pNLP-Mixer}: An Efficient all-{MLP} Architecture for Language}. In \bibinfo{booktitle}{\emph{Proceedings of the 61st Annual Meeting of the Association for Computational Linguistics}}. \bibinfo{publisher}{Association for Computational Linguistics}, \bibinfo{pages}{53--60}.
\newblock


\bibitem[Geurts et~al\mbox{.}(2006)]%
        {geurts2006extremely}
\bibfield{author}{\bibinfo{person}{Pierre Geurts}, \bibinfo{person}{Damien Ernst}, {and} \bibinfo{person}{Louis Wehenkel}.} \bibinfo{year}{2006}\natexlab{}.
\newblock \showarticletitle{Extremely randomized trees}.
\newblock \bibinfo{journal}{\emph{Machine learning}}  \bibinfo{volume}{63} (\bibinfo{year}{2006}), \bibinfo{pages}{3--42}.
\newblock


\bibitem[Gorishniy et~al\mbox{.}(2022)]%
        {gorishniy2022embeddings}
\bibfield{author}{\bibinfo{person}{Yury Gorishniy}, \bibinfo{person}{Ivan Rubachev}, {and} \bibinfo{person}{Artem Babenko}.} \bibinfo{year}{2022}\natexlab{}.
\newblock \showarticletitle{On embeddings for numerical features in tabular deep learning}. In \bibinfo{booktitle}{\emph{NeurIPS}}. \bibinfo{pages}{24991--25004}.
\newblock


\bibitem[Gorishniy et~al\mbox{.}(2021)]%
        {gorishniy2021revisiting}
\bibfield{author}{\bibinfo{person}{Yury Gorishniy}, \bibinfo{person}{Ivan Rubachev}, \bibinfo{person}{Valentin Khrulkov}, {and} \bibinfo{person}{Artem Babenko}.} \bibinfo{year}{2021}\natexlab{}.
\newblock \showarticletitle{Revisiting deep learning models for tabular data}. In \bibinfo{booktitle}{\emph{NeurIPS}}. \bibinfo{pages}{18932--18943}.
\newblock


\bibitem[Grinsztajn et~al\mbox{.}(2022)]%
        {grinsztajn2022tree}
\bibfield{author}{\bibinfo{person}{L{\'e}o Grinsztajn}, \bibinfo{person}{Edouard Oyallon}, {and} \bibinfo{person}{Ga{\"e}l Varoquaux}.} \bibinfo{year}{2022}\natexlab{}.
\newblock \showarticletitle{Why do tree-based models still outperform deep learning on typical tabular data?}. In \bibinfo{booktitle}{\emph{NeurIPS}}.
\newblock


\bibitem[Guo et~al\mbox{.}(2022)]%
        {guo2022hire}
\bibfield{author}{\bibinfo{person}{Jianyuan Guo}, \bibinfo{person}{Yehui Tang}, {et~al\mbox{.}}} \bibinfo{year}{2022}\natexlab{}.
\newblock \showarticletitle{{Hire-MLP: Vision MLP} via hierarchical rearrangement}. In \bibinfo{booktitle}{\emph{CVPR}}. \bibinfo{pages}{826--836}.
\newblock


\bibitem[He et~al\mbox{.}(2014)]%
        {he2014practical}
\bibfield{author}{\bibinfo{person}{Xinran He}, \bibinfo{person}{Junfeng Pan}, {et~al\mbox{.}}} \bibinfo{year}{2014}\natexlab{}.
\newblock \showarticletitle{Practical lessons from predicting clicks on ads at {Facebook}}. In \bibinfo{booktitle}{\emph{Proceedings of the International Workshop on Data Mining for Online Advertising}}.
\newblock


\bibitem[Hou et~al\mbox{.}(2020)]%
        {hou2020dynabert}
\bibfield{author}{\bibinfo{person}{Lu Hou}, \bibinfo{person}{Zhiqi Huang}, \bibinfo{person}{Lifeng Shang}, \bibinfo{person}{Xin Jiang}, \bibinfo{person}{Xiao Chen}, {and} \bibinfo{person}{Qun Liu}.} \bibinfo{year}{2020}\natexlab{}.
\newblock \showarticletitle{{DynaBERT: Dynamic BERT} with adaptive width and depth}. In \bibinfo{booktitle}{\emph{NeurIPS}}, Vol.~\bibinfo{volume}{33}. \bibinfo{pages}{9782--9793}.
\newblock


\bibitem[Howard and Gugger(2020)]%
        {howard2020fastai}
\bibfield{author}{\bibinfo{person}{Jeremy Howard} {and} \bibinfo{person}{Sylvain Gugger}.} \bibinfo{year}{2020}\natexlab{}.
\newblock \showarticletitle{Fastai: A layered API for deep learning}.
\newblock \bibinfo{journal}{\emph{Information}} \bibinfo{volume}{11}, \bibinfo{number}{2} (\bibinfo{year}{2020}), \bibinfo{pages}{108}.
\newblock


\bibitem[Huang et~al\mbox{.}(2020)]%
        {huang2020tabtransformer}
\bibfield{author}{\bibinfo{person}{Xin Huang}, \bibinfo{person}{Ashish Khetan}, \bibinfo{person}{Milan Cvitkovic}, {and} \bibinfo{person}{Zohar Karnin}.} \bibinfo{year}{2020}\natexlab{}.
\newblock \showarticletitle{{TabTransformer}: Tabular data modeling using contextual embeddings}.
\newblock \bibinfo{journal}{\emph{arXiv preprint arXiv:2012.06678}} (\bibinfo{year}{2020}).
\newblock


\bibitem[Kadra et~al\mbox{.}(2021)]%
        {kadra2021well}
\bibfield{author}{\bibinfo{person}{Arlind Kadra}, \bibinfo{person}{Marius Lindauer}, \bibinfo{person}{Frank Hutter}, {and} \bibinfo{person}{Josif Grabocka}.} \bibinfo{year}{2021}\natexlab{}.
\newblock \showarticletitle{Well-tuned simple nets excel on tabular datasets}. In \bibinfo{booktitle}{\emph{NeurIPS}}. \bibinfo{pages}{23928--23941}.
\newblock


\bibitem[Katzir et~al\mbox{.}(2020)]%
        {katzir2020net}
\bibfield{author}{\bibinfo{person}{Liran Katzir}, \bibinfo{person}{Gal Elidan}, {and} \bibinfo{person}{Ran El-Yaniv}.} \bibinfo{year}{2020}\natexlab{}.
\newblock \showarticletitle{{Net-DNF}: Effective deep modeling of tabular data}. In \bibinfo{booktitle}{\emph{ICLR}}.
\newblock


\bibitem[Ke et~al\mbox{.}(2017)]%
        {ke2017lightgbm}
\bibfield{author}{\bibinfo{person}{Guolin Ke}, \bibinfo{person}{Qi Meng}, \bibinfo{person}{Thomas Finley}, \bibinfo{person}{Taifeng Wang}, \bibinfo{person}{Wei Chen}, \bibinfo{person}{Weidong Ma}, \bibinfo{person}{Qiwei Ye}, {and} \bibinfo{person}{Tie-Yan Liu}.} \bibinfo{year}{2017}\natexlab{}.
\newblock \showarticletitle{{LightGBM}: A highly efficient gradient boosting decision tree}. In \bibinfo{booktitle}{\emph{NeurIPS}}.
\newblock


\bibitem[Kenton and Toutanova(2019)]%
        {kenton2019bert}
\bibfield{author}{\bibinfo{person}{Jacob Devlin Ming-Wei~Chang Kenton} {and} \bibinfo{person}{Lee~Kristina Toutanova}.} \bibinfo{year}{2019}\natexlab{}.
\newblock \showarticletitle{{BERT}: Pre-training of Deep Bidirectional {Transformers} for Language Understanding}. In \bibinfo{booktitle}{\emph{NAACL-HLT}}. \bibinfo{pages}{4171--4186}.
\newblock


\bibitem[Klambauer et~al\mbox{.}(2017)]%
        {klambauer2017self}
\bibfield{author}{\bibinfo{person}{G{\"u}nter Klambauer}, \bibinfo{person}{Thomas Unterthiner}, \bibinfo{person}{Andreas Mayr}, {and} \bibinfo{person}{Sepp Hochreiter}.} \bibinfo{year}{2017}\natexlab{}.
\newblock \showarticletitle{Self-normalizing neural networks}. In \bibinfo{booktitle}{\emph{NeurIPS}}, Vol.~\bibinfo{volume}{30}.
\newblock


\bibitem[Kohavi et~al\mbox{.}(1996)]%
        {kohavi1996scaling}
\bibfield{author}{\bibinfo{person}{Ron Kohavi} {et~al\mbox{.}}} \bibinfo{year}{1996}\natexlab{}.
\newblock \showarticletitle{Scaling up the accuracy of {Naive-Bayes} classifiers: A decision-tree hybrid}. In \bibinfo{booktitle}{\emph{KDD}}, Vol.~\bibinfo{volume}{96}. \bibinfo{pages}{202--207}.
\newblock


\bibitem[Li et~al\mbox{.}(1984)]%
        {li1984classification}
\bibfield{author}{\bibinfo{person}{Bin Li}, \bibinfo{person}{J Friedman}, \bibinfo{person}{R Olshen}, {and} \bibinfo{person}{C Stone}.} \bibinfo{year}{1984}\natexlab{}.
\newblock \showarticletitle{Classification and regression trees {(CART)}}.
\newblock \bibinfo{journal}{\emph{Biometrics}} (\bibinfo{year}{1984}).
\newblock


\bibitem[Liu et~al\mbox{.}(2021)]%
        {liu2021pay}
\bibfield{author}{\bibinfo{person}{Hanxiao Liu}, \bibinfo{person}{Zihang Dai}, \bibinfo{person}{David So}, {and} \bibinfo{person}{Quoc~V Le}.} \bibinfo{year}{2021}\natexlab{}.
\newblock \showarticletitle{Pay attention to {MLPs}}. In \bibinfo{booktitle}{\emph{NeurIPS}}. \bibinfo{pages}{9204--9215}.
\newblock


\bibitem[Louizos et~al\mbox{.}(2018)]%
        {louizos2018learning}
\bibfield{author}{\bibinfo{person}{Christos Louizos}, \bibinfo{person}{Max Welling}, {and} \bibinfo{person}{Diederik~P Kingma}.} \bibinfo{year}{2018}\natexlab{}.
\newblock \showarticletitle{Learning Sparse Neural Networks through {L\_0} Regularization}. In \bibinfo{booktitle}{\emph{ICLR}}.
\newblock


\bibitem[Ma et~al\mbox{.}(2023)]%
        {ma2023llm}
\bibfield{author}{\bibinfo{person}{Xinyin Ma}, \bibinfo{person}{Gongfan Fang}, {and} \bibinfo{person}{Xinchao Wang}.} \bibinfo{year}{2023}\natexlab{}.
\newblock \showarticletitle{{LLM-Pruner}: On the Structural Pruning of Large Language Models}. In \bibinfo{booktitle}{\emph{NeurIPS}}.
\newblock


\bibitem[Mikolov et~al\mbox{.}(2013)]%
        {mikolov2013efficient}
\bibfield{author}{\bibinfo{person}{Tomas Mikolov}, \bibinfo{person}{Kai Chen}, {et~al\mbox{.}}} \bibinfo{year}{2013}\natexlab{}.
\newblock \showarticletitle{Efficient estimation of word representations in vector space}.
\newblock \bibinfo{journal}{\emph{arXiv preprint arXiv:1301.3781}} (\bibinfo{year}{2013}).
\newblock


\bibitem[Pace and Barry(1997)]%
        {pace1997sparse}
\bibfield{author}{\bibinfo{person}{R~Kelley Pace} {and} \bibinfo{person}{Ronald Barry}.} \bibinfo{year}{1997}\natexlab{}.
\newblock \showarticletitle{Sparse spatial autoregressions}.
\newblock \bibinfo{journal}{\emph{Statistics \& Probability Letters}} \bibinfo{volume}{33}, \bibinfo{number}{3} (\bibinfo{year}{1997}), \bibinfo{pages}{291--297}.
\newblock


\bibitem[Pedregosa et~al\mbox{.}(2011)]%
        {scikit-learn}
\bibfield{author}{\bibinfo{person}{F. Pedregosa}, \bibinfo{person}{G. Varoquaux}, {et~al\mbox{.}}} \bibinfo{year}{2011}\natexlab{}.
\newblock \showarticletitle{Scikit-learn: Machine Learning in {P}ython}.
\newblock \bibinfo{journal}{\emph{Journal of Machine Learning Research}}  \bibinfo{volume}{12} (\bibinfo{year}{2011}), \bibinfo{pages}{2825--2830}.
\newblock


\bibitem[Popov et~al\mbox{.}(2019)]%
        {popov2019neural}
\bibfield{author}{\bibinfo{person}{Sergei Popov}, \bibinfo{person}{Stanislav Morozov}, {and} \bibinfo{person}{Artem Babenko}.} \bibinfo{year}{2019}\natexlab{}.
\newblock \showarticletitle{Neural Oblivious Decision Ensembles for Deep Learning on Tabular Data}. In \bibinfo{booktitle}{\emph{ICLR}}.
\newblock


\bibitem[Prokhorenkova et~al\mbox{.}(2018)]%
        {prokhorenkova2018catboost}
\bibfield{author}{\bibinfo{person}{Liudmila Prokhorenkova}, \bibinfo{person}{Gleb Gusev}, \bibinfo{person}{Aleksandr Vorobev}, \bibinfo{person}{Anna~Veronika Dorogush}, {and} \bibinfo{person}{Andrey Gulin}.} \bibinfo{year}{2018}\natexlab{}.
\newblock \showarticletitle{{CatBoost}: Unbiased boosting with categorical features}. In \bibinfo{booktitle}{\emph{NeurIPS}}.
\newblock


\bibitem[Radford et~al\mbox{.}(2021)]%
        {radford2021learning}
\bibfield{author}{\bibinfo{person}{Alec Radford}, \bibinfo{person}{Jong~Wook Kim}, {et~al\mbox{.}}} \bibinfo{year}{2021}\natexlab{}.
\newblock \showarticletitle{Learning transferable visual models from natural language supervision}. In \bibinfo{booktitle}{\emph{ICML}}. \bibinfo{pages}{8748--8763}.
\newblock


\bibitem[Ruiz et~al\mbox{.}(2023)]%
        {ruiz2023enabling}
\bibfield{author}{\bibinfo{person}{Camilo Ruiz}, \bibinfo{person}{Hongyu Ren}, \bibinfo{person}{Kexin Huang}, {and} \bibinfo{person}{Jure Leskovec}.} \bibinfo{year}{2023}\natexlab{}.
\newblock \showarticletitle{Enabling tabular deep learning when $ d \gg n $ with an auxiliary knowledge graph}.
\newblock \bibinfo{journal}{\emph{arXiv preprint arXiv:2306.04766}} (\bibinfo{year}{2023}).
\newblock


\bibitem[Seo et~al\mbox{.}(2017)]%
        {seo2017interpretable}
\bibfield{author}{\bibinfo{person}{Sungyong Seo}, \bibinfo{person}{Jing Huang}, \bibinfo{person}{Hao Yang}, {and} \bibinfo{person}{Yan Liu}.} \bibinfo{year}{2017}\natexlab{}.
\newblock \showarticletitle{Interpretable convolutional neural networks with dual local and global attention for review rating prediction}. In \bibinfo{booktitle}{\emph{Proceedings of the 11th ACM Conference on Recommender Systems}}. \bibinfo{pages}{297--305}.
\newblock


\bibitem[Shwartz-Ziv and Armon(2022)]%
        {shwartz2022tabular}
\bibfield{author}{\bibinfo{person}{Ravid Shwartz-Ziv} {and} \bibinfo{person}{Amitai Armon}.} \bibinfo{year}{2022}\natexlab{}.
\newblock \showarticletitle{Tabular data: Deep learning is not all you need}.
\newblock \bibinfo{journal}{\emph{Information Fusion}}  \bibinfo{volume}{81} (\bibinfo{year}{2022}), \bibinfo{pages}{84--90}.
\newblock


\bibitem[Somepalli et~al\mbox{.}(2022)]%
        {somepalli2022saint}
\bibfield{author}{\bibinfo{person}{Gowthami Somepalli}, \bibinfo{person}{Avi Schwarzschild}, \bibinfo{person}{Micah Goldblum}, \bibinfo{person}{C~Bayan Bruss}, {and} \bibinfo{person}{Tom Goldstein}.} \bibinfo{year}{2022}\natexlab{}.
\newblock \showarticletitle{{SAINT}: Improved Neural Networks for Tabular Data via Row Attention and Contrastive Pre-Training}. In \bibinfo{booktitle}{\emph{NeurIPS 2022 First Table Representation Workshop}}.
\newblock


\bibitem[Song et~al\mbox{.}(2019)]%
        {song2019autoint}
\bibfield{author}{\bibinfo{person}{Weiping Song}, \bibinfo{person}{Chence Shi}, \bibinfo{person}{Zhiping Xiao}, \bibinfo{person}{Zhijian Duan}, \bibinfo{person}{Yewen Xu}, \bibinfo{person}{Ming Zhang}, {and} \bibinfo{person}{Jian Tang}.} \bibinfo{year}{2019}\natexlab{}.
\newblock \showarticletitle{{AutoInt}: Automatic feature interaction learning via self-attentive neural networks}. In \bibinfo{booktitle}{\emph{CIKM}}. \bibinfo{pages}{1161--1170}.
\newblock


\bibitem[Sun et~al\mbox{.}(2023)]%
        {sun2023simple}
\bibfield{author}{\bibinfo{person}{Mingjie Sun}, \bibinfo{person}{Zhuang Liu}, \bibinfo{person}{Anna Bair}, {and} \bibinfo{person}{J~Zico Kolter}.} \bibinfo{year}{2023}\natexlab{}.
\newblock \showarticletitle{A Simple and Effective Pruning Approach for Large Language Models}.
\newblock \bibinfo{journal}{\emph{arXiv preprint arXiv:2306.11695}} (\bibinfo{year}{2023}).
\newblock


\bibitem[Tang et~al\mbox{.}(2022)]%
        {tang2022sparse}
\bibfield{author}{\bibinfo{person}{Chuanxin Tang}, \bibinfo{person}{Yucheng Zhao}, {et~al\mbox{.}}} \bibinfo{year}{2022}\natexlab{}.
\newblock \showarticletitle{Sparse {MLP} for image recognition: Is self-attention really necessary?}. In \bibinfo{booktitle}{\emph{AAAI}}. \bibinfo{pages}{2344--2351}.
\newblock


\bibitem[Tarvainen and Valpola(2017)]%
        {tarvainen2017mean}
\bibfield{author}{\bibinfo{person}{Antti Tarvainen} {and} \bibinfo{person}{Harri Valpola}.} \bibinfo{year}{2017}\natexlab{}.
\newblock \showarticletitle{Mean teachers are better role models: Weight-averaged consistency targets improve semi-supervised deep learning results}. In \bibinfo{booktitle}{\emph{NeurIPS}}, Vol.~\bibinfo{volume}{30}.
\newblock


\bibitem[Tolstikhin et~al\mbox{.}(2021)]%
        {tolstikhin2021mlp}
\bibfield{author}{\bibinfo{person}{Ilya~O Tolstikhin}, \bibinfo{person}{Neil Houlsby}, \bibinfo{person}{Alexander Kolesnikov}, \bibinfo{person}{Lucas Beyer}, \bibinfo{person}{Xiaohua Zhai}, \bibinfo{person}{Thomas Unterthiner}, \bibinfo{person}{Jessica Yung}, \bibinfo{person}{Andreas Steiner}, \bibinfo{person}{Daniel Keysers}, \bibinfo{person}{Jakob Uszkoreit}, {et~al\mbox{.}}} \bibinfo{year}{2021}\natexlab{}.
\newblock \showarticletitle{{MLP-Mixer: An all-MLP} architecture for vision}. In \bibinfo{booktitle}{\emph{NeurIPS}}. \bibinfo{pages}{24261--24272}.
\newblock


\bibitem[Tu et~al\mbox{.}(2022)]%
        {tu2022maxim}
\bibfield{author}{\bibinfo{person}{Zhengzhong Tu}, \bibinfo{person}{Hossein Talebi}, \bibinfo{person}{Han Zhang}, \bibinfo{person}{Feng Yang}, \bibinfo{person}{Peyman Milanfar}, \bibinfo{person}{Alan Bovik}, {and} \bibinfo{person}{Yinxiao Li}.} \bibinfo{year}{2022}\natexlab{}.
\newblock \showarticletitle{{MAXIM}: Multi-Axis {MLP} for image processing}. In \bibinfo{booktitle}{\emph{CVPR}}. \bibinfo{pages}{5769--5780}.
\newblock


\bibitem[Uddin et~al\mbox{.}(2019)]%
        {uddin2019comparing}
\bibfield{author}{\bibinfo{person}{Shahadat Uddin}, \bibinfo{person}{Arif Khan}, \bibinfo{person}{Md~Ekramul Hossain}, {and} \bibinfo{person}{Mohammad~Ali Moni}.} \bibinfo{year}{2019}\natexlab{}.
\newblock \showarticletitle{Comparing different supervised machine learning algorithms for disease prediction}.
\newblock \bibinfo{journal}{\emph{BMC Medical Informatics and Decision Making}} (\bibinfo{year}{2019}), \bibinfo{pages}{1--16}.
\newblock


\bibitem[Vaswani et~al\mbox{.}(2017)]%
        {vaswani2017attention}
\bibfield{author}{\bibinfo{person}{Ashish Vaswani}, \bibinfo{person}{Noam Shazeer}, \bibinfo{person}{Niki Parmar}, \bibinfo{person}{Jakob Uszkoreit}, \bibinfo{person}{Llion Jones}, \bibinfo{person}{Aidan~N Gomez}, \bibinfo{person}{{\L}ukasz Kaiser}, {and} \bibinfo{person}{Illia Polosukhin}.} \bibinfo{year}{2017}\natexlab{}.
\newblock \showarticletitle{Attention is all you need}. In \bibinfo{booktitle}{\emph{NeurIPS}}.
\newblock


\bibitem[Wang et~al\mbox{.}(2021)]%
        {wang2021dcn}
\bibfield{author}{\bibinfo{person}{Ruoxi Wang}, \bibinfo{person}{Rakesh Shivanna}, \bibinfo{person}{Derek Cheng}, \bibinfo{person}{Sagar Jain}, \bibinfo{person}{Dong Lin}, \bibinfo{person}{Lichan Hong}, {and} \bibinfo{person}{Ed Chi}.} \bibinfo{year}{2021}\natexlab{}.
\newblock \showarticletitle{{DCN V2}: Improved deep \& cross network and practical lessons for web-scale learning to rank systems}. In \bibinfo{booktitle}{\emph{WWW}}. \bibinfo{pages}{1785--1797}.
\newblock


\bibitem[Wang and Sun(2022)]%
        {wang2022transtab}
\bibfield{author}{\bibinfo{person}{Zifeng Wang} {and} \bibinfo{person}{Jimeng Sun}.} \bibinfo{year}{2022}\natexlab{}.
\newblock \showarticletitle{{TransTab}: Learning transferable tabular {Transformers} across tables}. In \bibinfo{booktitle}{\emph{NeurIPS}}, Vol.~\bibinfo{volume}{35}. \bibinfo{pages}{2902--2915}.
\newblock


\bibitem[Wang et~al\mbox{.}(2020)]%
        {wang2020structured}
\bibfield{author}{\bibinfo{person}{Ziheng Wang}, \bibinfo{person}{Jeremy Wohlwend}, {and} \bibinfo{person}{Tao Lei}.} \bibinfo{year}{2020}\natexlab{}.
\newblock \showarticletitle{Structured Pruning of Large Language Models}. In \bibinfo{booktitle}{\emph{EMNLP}}. \bibinfo{pages}{6151--6162}.
\newblock


\bibitem[Wortsman et~al\mbox{.}(2022)]%
        {wortsman2022model}
\bibfield{author}{\bibinfo{person}{Mitchell Wortsman}, \bibinfo{person}{Gabriel Ilharco}, \bibinfo{person}{Samir~Ya Gadre}, \bibinfo{person}{Rebecca Roelofs}, \bibinfo{person}{Raphael Gontijo-Lopes}, \bibinfo{person}{Ari~S Morcos}, \bibinfo{person}{Hongseok Namkoong}, \bibinfo{person}{Ali Farhadi}, \bibinfo{person}{Yair Carmon}, \bibinfo{person}{Simon Kornblith}, {et~al\mbox{.}}} \bibinfo{year}{2022}\natexlab{}.
\newblock \showarticletitle{Model soups: Averaging weights of multiple fine-tuned models improves accuracy without increasing inference time}. In \bibinfo{booktitle}{\emph{ICML}}. \bibinfo{pages}{23965--23998}.
\newblock


\bibitem[Xia et~al\mbox{.}(2022)]%
        {xia2022structured}
\bibfield{author}{\bibinfo{person}{Mengzhou Xia}, \bibinfo{person}{Zexuan Zhong}, {and} \bibinfo{person}{Danqi Chen}.} \bibinfo{year}{2022}\natexlab{}.
\newblock \showarticletitle{Structured Pruning Learns Compact and Accurate Models}. In \bibinfo{booktitle}{\emph{ACL}}.
\newblock


\bibitem[Yan et~al\mbox{.}(2023)]%
        {yan2023t2g}
\bibfield{author}{\bibinfo{person}{Jiahuan Yan}, \bibinfo{person}{Jintai Chen}, \bibinfo{person}{Yixuan Wu}, \bibinfo{person}{Danny~Z Chen}, {and} \bibinfo{person}{Jian Wu}.} \bibinfo{year}{2023}\natexlab{}.
\newblock \showarticletitle{{T2G-Former}: Organizing tabular features into relation graphs promotes heterogeneous feature interaction}. In \bibinfo{booktitle}{\emph{AAAI}}.
\newblock


\bibitem[Yan et~al\mbox{.}(2024)]%
        {yan2024making}
\bibfield{author}{\bibinfo{person}{Jiahuan Yan}, \bibinfo{person}{Bo Zheng}, \bibinfo{person}{Hongxia Xu}, \bibinfo{person}{Yiheng Zhu}, \bibinfo{person}{Danny Chen}, \bibinfo{person}{Jimeng Sun}, \bibinfo{person}{Jian Wu}, {and} \bibinfo{person}{Jintai Chen}.} \bibinfo{year}{2024}\natexlab{}.
\newblock \showarticletitle{Making Pre-trained Language Models Great on Tabular Prediction}. In \bibinfo{booktitle}{\emph{ICLR}}.
\newblock


\bibitem[Yang et~al\mbox{.}(2022)]%
        {yang2022locally}
\bibfield{author}{\bibinfo{person}{Junchen Yang}, \bibinfo{person}{Ofir Lindenbaum}, {and} \bibinfo{person}{Yuval Kluger}.} \bibinfo{year}{2022}\natexlab{}.
\newblock \showarticletitle{Locally sparse neural networks for tabular biomedical data}. In \bibinfo{booktitle}{\emph{ICML}}. PMLR, \bibinfo{pages}{25123--25153}.
\newblock


\bibitem[Zhang and Honavar(2003)]%
        {zhang2003learning}
\bibfield{author}{\bibinfo{person}{Jun Zhang} {and} \bibinfo{person}{Vasant Honavar}.} \bibinfo{year}{2003}\natexlab{}.
\newblock \showarticletitle{Learning from attribute value taxonomies and partially specified instances}. In \bibinfo{booktitle}{\emph{ICML}}.
\newblock


\bibitem[Zhang et~al\mbox{.}(2006)]%
        {zhang2006learning}
\bibfield{author}{\bibinfo{person}{Jun Zhang}, \bibinfo{person}{D-K Kang}, {et~al\mbox{.}}} \bibinfo{year}{2006}\natexlab{}.
\newblock \showarticletitle{Learning accurate and concise {Na{\"\i}ve Bayes} classifiers from attribute value taxonomies and data}.
\newblock \bibinfo{journal}{\emph{Knowledge and Information Systems}} (\bibinfo{year}{2006}).
\newblock


\bibitem[Zhang et~al\mbox{.}(2023)]%
        {zhang2023generative}
\bibfield{author}{\bibinfo{person}{Tianping Zhang}, \bibinfo{person}{Shaowen Wang}, \bibinfo{person}{Shuicheng Yan}, \bibinfo{person}{Jian Li}, {and} \bibinfo{person}{Qian Liu}.} \bibinfo{year}{2023}\natexlab{}.
\newblock \showarticletitle{Generative Table Pre-training Empowers Models for Tabular Prediction}.
\newblock \bibinfo{journal}{\emph{arXiv preprint arXiv:2305.09696}} (\bibinfo{year}{2023}).
\newblock


\bibitem[Zhao et~al\mbox{.}(2023)]%
        {zhao2023survey}
\bibfield{author}{\bibinfo{person}{Wayne~Xin Zhao}, \bibinfo{person}{Kun Zhou}, \bibinfo{person}{Junyi Li}, \bibinfo{person}{Tianyi Tang}, \bibinfo{person}{Xiaolei Wang}, \bibinfo{person}{Yupeng Hou}, \bibinfo{person}{Yingqian Min}, \bibinfo{person}{Beichen Zhang}, \bibinfo{person}{Junjie Zhang}, \bibinfo{person}{Zican Dong}, {et~al\mbox{.}}} \bibinfo{year}{2023}\natexlab{}.
\newblock \showarticletitle{A survey of large language models}.
\newblock \bibinfo{journal}{\emph{arXiv preprint arXiv:2303.18223}} (\bibinfo{year}{2023}).
\newblock


\bibitem[Zhu et~al\mbox{.}(2023)]%
        {zhu2023xtab}
\bibfield{author}{\bibinfo{person}{Bingzhao Zhu}, \bibinfo{person}{Xingjian Shi}, \bibinfo{person}{Nick Erickson}, \bibinfo{person}{Mu Li}, \bibinfo{person}{George Karypis}, {and} \bibinfo{person}{Mahsa Shoaran}.} \bibinfo{year}{2023}\natexlab{}.
\newblock \showarticletitle{{XTab}: Cross-table Pretraining for Tabular {Transformers}}. In \bibinfo{booktitle}{\emph{ICML}}.
\newblock


\end{thebibliography}

\balance
\appendix

\section{Benchmark Characteristics} \label{benchmark-information}

We provide detailed dataset statistical information of each benchmark in Table~\ref{app:bench-info}. These benchmarks exhibit broad data diversity in data scales and task types. From the FT-T benchmark to TabBen, the overall data volume is gradually reduced. We additionally visualize the respective winning rates of GBDT and DNN frameworks in Fig.~\ref{win-rate}, indicating varying framework preferences among the dataset collections used in different tabular prediction tasks. FT-T does not include GBDT baselines in its main benchmark, but has the most extremely large datasets. Overall, the FT-T benchmark is the extremely large-scale data collection (in both data volume and feature width), the T2G benchmark is a large one, the SAINT benchmark contains diverse data scales, and TabBen focuses on middle-size typical tables.

\section{Baseline Information} \label{baseline-info}

We list all the compared baselines in this section.
\begin{itemize}
    \item MLP: Vanilla multi-layer perception with no feature interaction.
    \item ResNet: A popular DNN backbone in vision applications.
    \item SNN~\cite{klambauer2017self}: An MLP-like architecture with SELU activation.
    \item GrowNet~\cite{badirli2020gradient}: MLPs built in a gradient boosted manner.
    \item NODE~\cite{popov2019neural}: Generalized oblivious decision tree ensembles.
    \item TabNet~\cite{arik2021tabnet}: A Transformer-based recurrent architecture emulating tree-based learning process.
    \item AutoInt~\cite{song2019autoint}: Attention-based feature embeddings.
    \item DCNv2~\cite{wang2021dcn}: An MLP-based architecture with the feature-crossing module.
    \item TabTransformer~\cite{huang2020tabtransformer}: A Transformer model concatenating numerical features and encoded categorical features.
    \item DANets~\cite{chen2022danets}: An MLP-based architecture with neural-guided feature selection and abstraction in each block.
    \item FT-Transformer~\cite{gorishniy2021revisiting}: A popular tabular Transformer encoding both numerical and categorical features.
    \item T2G-Former~\cite{yan2023t2g}: A tabular Transformer with automatic relation graph estimation for selective feature interaction.
    \item SAINT~\cite{somepalli2022saint}: A Transformer-like architecture performing row-level and column-level attention, and contrastively pre-training to minimize the differences between data points and their augmented views.
    \item XGBoost~\cite{chen2016xgboost}: A predominant GBDT implementation.
    \item CatBoost~\cite{prokhorenkova2018catboost}: A GBDT approach with oblivious decision trees.
    \item LightGBM~\cite{ke2017lightgbm}: An efficient GBDT implementation.
    \item RandomForest~\cite{breiman2001random}: A popular bagging ensemble algorithm of decision trees.
    \item ExtraTrees~\cite{geurts2006extremely}: A classical tree bagging implementation.
    \item k-NN~\cite{altman1992introduction}: Traditional supervised machine learning algorithms; two KNeighbors models are used (KNeighborsDist, KNeighborsUnif).
    \item NeuralNetFastAI~\cite{howard2020fastai}: FastAI neural network models that operate on tabular data.
    \item sklearn-GBDT~\cite{scikit-learn}: Two traditional GBDT implementations (GradientBoostingTree and HistGradientBoostingTrees) provided in the Scikit Learn package.
\end{itemize}

\section{Runtime Environment and Hyperparameters} \label{hyper-info}

\subsection{Runtime Environment}
All the experiments are conducted with PyTorch version 1.11.0, CUDA version 11.3, and Scikit Learn version 1.1.0, with each trial using an NVIDIA A100 PCIe 40GB and an Intel Xeon Processor 40C.

\subsection{Hyperparameters of T-MLP}
In the main experiments, we uniformly set the hidden size $d$ to 1024, the intermediate size $d'$ to 676 (2/3 of the hidden size), the sparsity rate to 0.33, and the residual dropout rate to 0.1, with three basic blocks for multi-class classification or extremely large binary classification datasets, and one block for the others. The learning rate of the single T-MLP is 1e-4, and the learning rates of the three branches in T-MLP ensemble are 1e-4, 5e-4, and 1e-3, respectively.

\subsection{Hyperparameters of Baselines}
For all the baselines on the FT-T and T2G benchmarks, we follow the given hyperparameter spaces and iteration times from the original benchmark papers to estimate the training costs.

% detailed results
\begin{table*}
\caption{AUC scores (the higher the better) of the baselines on the binary classification datasets in the SAINT benchmark.}
\label{app:saint-bin}
\centering
\begin{tabular}{@{}l|ccccccccc@{}}
\toprule
OpenML ID:     & 31    & 44    & 1017  & 1111  & 1487  & 1494  & 1590  & 4134  & 42178 \\ \midrule
RF   & 0.778 & 0.986 & 0.798 & 0.774 & 0.910 & 0.928 & 0.908 & 0.868 & 0.840 \\
ExtraTrees     & 0.764 & 0.986 & 0.811 & 0.748 & 0.921 & 0.935 & 0.903 & 0.856 & 0.831 \\
KNeighborsDist & 0.501 & 0.873 & 0.722 & 0.517 & 0.741 & 0.868 & 0.684 & 0.808 & 0.755 \\
KNeighborsUnif & 0.489 & 0.847 & 0.712 & 0.516 & 0.734 & 0.865 & 0.669 & 0.790 & 0.764 \\
LightGBM       & 0.752 & 0.989 & 0.829 & 0.815 & 0.919 & 0.923 & 0.930 & 0.860 & 0.854 \\
XGBoost        & 0.778 & 0.989 & 0.821 & 0.818 & 0.919 & 0.926 & 0.931 & 0.864 & 0.856 \\
CatBoost       & 0.788 & 0.988 & 0.838 & 0.818 & 0.917 & 0.937 & 0.930 & 0.862 & 0.841 \\
MLP            & 0.705 & 0.980 & 0.745 & 0.709 & 0.913 & 0.932 & 0.910 & 0.818 & 0.841 \\
TabNet         & 0.736 & 0.979 & 0.422 & 0.718 & 0.625 & 0.677 & 0.917 & 0.701 & 0.830 \\
TabTransformer & 0.771 & 0.982 & 0.729 & 0.763 & 0.884 & 0.913 & 0.907 & 0.809 & 0.841 \\
SAINT-s        & 0.774 & 0.982 & 0.781 & 0.804 & 0.906 & 0.933 & 0.922 & 0.819 & 0.858 \\
SAINT-i        & 0.774 & 0.981 & 0.759 & 0.816 & 0.920 & 0.934 & 0.919 & 0.845 & 0.854 \\
SAINT          & 0.790 & 0.991 & 0.843 & 0.808 & 0.919 & 0.937 & 0.921 & 0.853 & 0.857 \\ \midrule
T-MLP           & 0.805 & 0.983 & 0.818 & 0.814 & 0.924 & 0.933 & 0.924 & 0.853 & 0.862 \\
T-MLP(3)        & 0.802 & 0.983 & 0.821 & 0.816 & 0.924 & 0.935 & 0.925 & 0.855 & 0.861 \\ \bottomrule
\end{tabular}
\end{table*}

\begin{table*}
\caption{Accuracy scores (the higher the better) of the baselines on the multi-class classification datasets in the SAINT benchmark.}
\label{app:saint-mul}
\centering
\begin{tabular}{@{}l|ccccccc@{}}
\toprule
OpenML ID:     & 188   & 1596  & 4541  & 40685 & 41166 & 41169 & 42734 \\ \midrule
RF  & 0.653 & 0.953 & 0.607 & 0.999 & 0.671 & 0.358 & 0.743 \\
ExtraTrees     & 0.653 & 0.946 & 0.595 & 0.999 & 0.648 & 0.341 & 0.736 \\
KNeighborsDist & 0.442 & 0.965 & 0.491 & 0.997 & 0.620 & 0.205 & 0.685 \\
KNeighborsUnif & 0.422 & 0.963 & 0.489 & 0.997 & 0.605 & 0.189 & 0.693 \\
LightGBM       & 0.667 & 0.969 & 0.611 & 0.999 & 0.721 & 0.356 & 0.754 \\
XGBoost        & 0.612 & 0.928 & 0.611 & 0.999 & 0.707 & 0.356 & 0.752 \\
CatBoost       & 0.667 & 0.871 & 0.604 & 0.999 & 0.692 & 0.376 & 0.747 \\
MLP            & 0.388 & 0.915 & 0.597 & 0.997 & 0.707 & 0.378 & 0.733 \\
TabNet         & 0.259 & 0.744 & 0.517 & 0.997 & 0.599 & 0.243 & 0.630 \\
TabTransformer & 0.660 & 0.715 & 0.601 & 0.999 & 0.531 & 0.352 & 0.744 \\
SAINT-s        & 0.680 & 0.735 & 0.607 & 0.999 & 0.582 & 0.194 & 0.755 \\
SAINT-i        & 0.646 & 0.937 & 0.598 & 0.999 & 0.713 & 0.380 & 0.747 \\
SAINT          & 0.680 & 0.946 & 0.606 & 0.999 & 0.701 & 0.377 & 0.752 \\ \midrule
T-MLP           & 0.660 & 0.968 & 0.598 & 1.000 & 0.718 & 0.382 & 0.747 \\
T-MLP(3)        & 0.674 & 0.970 & 0.601 & 1.000 & 0.728 & 0.384 & 0.750 \\ \bottomrule
\end{tabular}
\end{table*}

\begin{table*}
\caption{RMSE scores (the lower the better) of the baselines on the regression datasets in the SAINT benchmark.}
\label{app:saint-reg}
\centering
\begin{tabular}{@{}l|cccccccccc@{}}
\toprule
OpenML ID:     & 422   & 541    & 42563      & 42571    & 42705  & 42724     & 42726 & 42727 & 42728  & 42729 \\ \midrule
RF    & 0.027 & 17.814 & 37085.577  & 1999.442 & 16.729 & 12375.312 & 2.476 & 0.149 & 13.700 & 1.767 \\
ExtraTrees      & 0.027 & 19.269 & 35049.267  & 1961.928 & 15.349 & 12505.090 & 2.522 & 0.147 & 13.578 & 1.849 \\
KNeighborsDist  & 0.029 & 25.054 & 46331.144  & 2617.202 & 14.496 & 13046.090 & 2.501 & 0.167 & 13.692 & 2.100 \\
KNeighborsUnif  & 0.029 & 24.698 & 47201.343  & 2629.277 & 18.397 & 12857.449 & 2.592 & 0.169 & 13.703 & 2.109 \\
LightGBM        & 0.027 & 19.871 & 32870.697  & 1898.032 & 13.018 & 11639.594 & 2.451 & 0.144 & 13.468 & 1.958 \\
XGBoost         & 0.028 & 13.791 & 36375.583  & 1903.027 & 12.311 & 11931.233 & 2.452 & 0.145 & 13.480 & 1.849 \\
CatBoost        & 0.027 & 14.060 & 35187.381  & 1886.593 & 11.890 & 11614.567 & 2.405 & 0.142 & 13.441 & 1.883 \\
NeuralNetFastAl & 0.028 & 22.756 & 42751.432  & 1991.774 & 15.892 & 11618.684 & 2.500 & 0.162 & 13.781 & 3.351 \\
TabNet          & 0.028 & 22.731 & 200802.769 & 1943.091 & 11.084 & 11613.275 & 2.175 & 0.183 & 16.665 & 2.310 \\
TabTransformer  & 0.028 & 21.600 & 37057.686  & 1980.696 & 15.693 & 11618.356 & 2.494 & 0.162 & 12.982 & 3.259 \\
SAINT--s        & 0.027 & 9.613  & 193430.703 & 1937.189 & 10.034 & 11580.835 & 2.145 & 0.158 & 12.603 & 1.833 \\
SAINT-i         & 0.028 & 12.564 & 33992.508  & 1997.111 & 11.513 & 11612.084 & 2.104 & 0.153 & 12.534 & 1.867 \\
SAINT           & 0.027 & 11.661 & 33112.387  & 1953.391 & 10.282 & 11577.678 & 2.113 & 0.145 & 12.578 & 1.882 \\ \midrule
T-MLP            & 0.027 & 11.643 & 21773.233  & 1946.203 & 9.027  & 11828.872 & 2.041 & 0.161 & 13.271 & 1.843 \\
T-MLP(3)         & 0.027 & 13.790 & 22185.024  & 1939.557 & 8.972  & 11762.376 & 2.049 & 0.161 & 13.016 & 1.852 \\ \bottomrule
\end{tabular}
\end{table*}

\begin{table*}
\caption{Accuracy scores (the higher the better) of the baselines for the binary classification tasks on the TabBen numerical datasets.}
\label{app:tabben-bin-num}
\centering
\begin{tabular}{@{}l|ccccccccc@{}}
\toprule
                     & eye   & MiniBooNE & Higgs & bank-market & covertype & MagicTele. & electricity & credit & jannis \\ \midrule
Resnet               & 0.574 & 0.937     & 0.694 & 0.794       & 0.803     & 0.858      & 0.809       & 0.761  & 0.746  \\
FT-T       & 0.586 & 0.937     & 0.706 & 0.804       & 0.813     & 0.851      & 0.820       & 0.765  & 0.766  \\
SAINT                & 0.589 & 0.935     & 0.707 & 0.791       & 0.803     & 0.851      & 0.818       & 0.760  & 0.773  \\
GBT & 0.637 & 0.932     & 0.711 & 0.803       & 0.819     & 0.851      & 0.862       & 0.772  & 0.770  \\
XGBoost              & 0.655 & 0.936     & 0.714 & 0.804       & 0.819     & 0.859      & 0.868       & 0.774  & 0.778  \\
RF         & 0.650 & 0.927     & 0.708 & 0.798       & 0.827     & 0.853      & 0.861       & 0.772  & 0.773  \\
MLP                  & 0.569 & 0.935     & 0.689 & 0.792       & 0.789     & 0.847      & 0.810       & 0.760  & 0.746  \\ \midrule
T-MLP                 & 0.610 & 0.946     & 0.731 & 0.802       & 0.909     & 0.859      & 0.842       & 0.772  & 0.800  \\
T-MLP(3)              & 0.613 & 0.947     & 0.733 & 0.803       & 0.915     & 0.861      & 0.848       & 0.775  & 0.799 \\ \bottomrule
\end{tabular}
\end{table*}

\begin{table*}
\caption{R-Squared scores (the higher the better) of the baselines for the regression tasks on the TabBen numerical datasets.}
\label{app:tabben-reg-num}
\centering
\setlength{\tabcolsep}{2.0pt}
\begin{tabular}{@{}l|cccccccccccccc@{}}
\toprule
                     & elevators & Bike  & houses & nyc-taxi & pol   & sulfur & Ailerons & wine  & supercon. & house sales & Brazilian & Miami & cpu act & diamonds \\ \midrule
Resnet               & 0.910     & 0.669 & 0.821  & 0.468    & 0.948 & 0.819  & 0.835    & 0.363 & 0.895     & 0.868       & 0.998     & 0.914      & 0.982   & 0.942    \\
FT-T       & 0.914     & 0.671 & 0.832  & 0.476    & 0.995 & 0.838  & 0.844    & 0.359 & 0.885     & 0.875       & 0.998     & 0.919      & 0.978   & 0.944    \\
SAINT                & 0.923     & 0.684 & 0.820  & 0.496    & 0.996 & 0.788  & 0.784    & 0.374 & 0.894     & 0.879       & 0.994     & 0.921      & 0.984   & 0.944    \\
GBT & 0.863     & 0.690 & 0.840  & 0.554    & 0.979 & 0.806  & 0.843    & 0.458 & 0.905     & 0.884       & 0.995     & 0.924      & 0.985   & 0.945    \\
XGBoost              & 0.908     & 0.695 & 0.852  & 0.553    & 0.990 & 0.865  & 0.844    & 0.498 & 0.911     & 0.887       & 0.998     & 0.936      & 0.986   & 0.946    \\
RF         & 0.841     & 0.687 & 0.829  & 0.563    & 0.989 & 0.859  & 0.839    & 0.504 & 0.909     & 0.871       & 0.993     & 0.924      & 0.983   & 0.945    \\ \midrule
T-MLP                 & 0.875     & 0.694 & 0.834  & 0.560    & 0.995 & 0.853  & 0.840    & 0.410 & 0.894     & 0.886       & 0.993     & 0.939      & 0.982   & 0.949    \\
T-MLP(3)              & 0.908     & 0.698 & 0.838  & 0.566    & 0.996 & 0.860  & 0.843    & 0.416 & 0.899     & 0.888       & 0.995     & 0.939      & 0.983   & 0.950   \\ \bottomrule
\end{tabular}
\end{table*}

\begin{table*}
\caption{Accuracy scores (the higher the better) of the baselines for the binary classification tasks on the TabBen categorical datasets.}
\label{app:tabben-bin-cat}
\centering
\begin{tabular}{@{}l|cccccc@{}}
\toprule
                         & eye   & road-safety & electricity & covertype & rl    & compass \\ \midrule
FT-T           & 0.598 & 0.767       & 0.842       & 0.867     & 0.703 & 0.753   \\
Resnet                   & 0.579 & 0.761       & 0.826       & 0.853     & 0.706 & 0.745   \\
SAINT                    & 0.585 & 0.764       & 0.834       & 0.850     & 0.682 & 0.719   \\
GBT     & 0.639 & 0.762       & 0.880       & 0.856     & 0.776 & 0.741   \\
XGBoost                  & 0.668 & 0.767       & 0.887       & 0.864     & 0.770 & 0.769   \\
HistGBT & 0.636 & 0.765       & 0.882       & 0.845     & 0.761 & 0.751   \\
RF             & 0.657 & 0.759       & 0.878       & 0.859     & 0.798 & 0.793   \\ \midrule
T-MLP                     & 0.605 & 0.786       & 0.880       & 0.882     & 0.757 & 0.785   \\
T-MLP(3)                  & 0.609 & 0.786       & 0.881       & 0.880     & 0.762 & 0.790  \\ \bottomrule
\end{tabular}
\end{table*}

\begin{table*}
\caption{R-Squared scores (the higher the better) of the baselines for the regression tasks on the TabBen categorical datasets.}
\label{app:tabben-reg-cat}
\centering
\begin{tabular}{@{}l|cccccccccc@{}}
\toprule
                         & Bike  & particulate & Brazilian & diamonds & black & nyc-taxi & analcatdata & OnlineNews & Mercedes & house sales \\ \midrule
FT-T           & 0.937 & 0.673       & 0.883     & 0.990    & 0.379 & 0.511    & 0.977       & 0.143      & 0.548    & 0.891       \\
Resnet                   & 0.936 & 0.658       & 0.878     & 0.989    & 0.360 & 0.451    & 0.978       & 0.130      & 0.545    & 0.881       \\
GBT     & 0.942 & 0.683       & 0.995     & 0.990    & 0.615 & 0.573    & 0.981       & 0.153      & 0.578    & 0.891       \\
XGBoost                  & 0.946 & 0.691       & 0.998     & 0.991    & 0.619 & 0.578    & 0.983       & 0.162      & 0.578    & 0.896       \\
HistGBT & 0.942 & 0.690       & 0.993     & 0.991    & 0.616 & 0.539    & 0.982       & 0.156      & 0.576    & 0.890       \\
RF             & 0.938 & 0.674       & 0.993     & 0.988    & 0.609 & 0.585    & 0.981       & 0.149      & 0.575    & 0.875       \\ \midrule
T-MLP                     & 0.938 & 0.692       & 0.996     & 0.990    & 0.620 & 0.571    & 0.990       & 0.154      & 0.576    & 0.893       \\
T-MLP(3)                  & 0.942 & 0.698       & 0.996     & 0.993    & 0.622 & 0.580    & 0.990       & 0.158      & 0.578    & 0.894      \\ \bottomrule
\end{tabular}
\end{table*}

\end{document}